\documentclass[runningheads]{llncs}
\pdfoutput=1

\usepackage[dvipsnames]{xcolor}
\usepackage{times}
\usepackage{epsfig}
\usepackage{graphicx}

\usepackage{amsmath}
\usepackage{amssymb}
\usepackage{newtxtt} 
\usepackage{longtable}
\usepackage{scalefnt}
\usepackage{pgfplots}
\usepackage{graphicx, nicefrac}
\usetikzlibrary{patterns,shapes.arrows}
\usetikzlibrary{shapes.symbols}
\pgfplotsset{compat=newest}
\usetikzlibrary{math}
\usetikzlibrary{fpu}
\usepgflibrary{fpu}
\usepackage{xargs} 
\usepackage{hyperref}
\usepackage{placeins} 

\pgfplotsset{compat=1.3}
\usepgfplotslibrary{external} 
\tikzsetexternalprefix{line_analysis/tikz_figures/}
{main_empirically_explaining_SGD}

\tikzset{external/mode=graphics if exists}

\newcommand{\pgfmathparseFPU}[1]{\begingroup%
	\pgfkeys{/pgf/fpu,/pgf/fpu/output format=fixed}%
	\pgfmathparse{#1}%
	\pgfmathsmuggle\pgfmathresult\endgroup}

\pgfmathdeclarefunction{symlog}{2}{\pgfmathparse{((#1>#2)-(#1<-#2))*(ln(max(abs(#1/#2),1))+1+((#1>=-#2)*(#1<=#2)*#1/#2)}}
\pgfmathdeclarefunction{symexp}{2}{\pgfmathparse{((#1>1)-(#1<-1))*#2*exp(abs(#1)-1)+((#1>=-1)*(#1<=1)*#2*#1)}}

\begin{document}
	\newcommand{\de}{\partial}
\newcommand{\dk}{\tensor[^{\de}]{k}{}}
\newcommand{\ddk}{\tensor[^{\de^2}]{k}{}}
\newcommand{\ddkd}{\tensor[^{\de^2}]{k}{^{\de}}}
\newcommand{\dddk}{\tensor[^{\de^3}]{k}{}}
\newcommand{\dddkd}{\tensor[^{\de^3}]{k}{^{\de}}}
\newcommand{\kd}{\tensor{k}{^{\de}}}
\newcommand{\dkd}{\tensor[^{\de}]{k}{^{\de}}}
\renewcommand{\vec}{\boldsymbol} 
\newcommand{\Id}{\mathbf{I}}
\newcommand{\Trans}{^\top}
\newcommand{\bl}{l_{\mathbb{B},t}}
\newcommand{\vbl}{\mathbf{l}_{\mathbb{B},t}}
\newcommand{\fbl}{l_t}
\newcommand{\bld}{l^\prime_{\mathbb{B},t}}
\newcommand{\vbld}{\mathbf{l}^\prime_{\mathbb{B},t}}
\newcommand{\fbld}{l^\prime_{t}}
\newcommand{\gp}{GP}
\newcommand{\pW}{p^\text{Wolfe}}
\newcommand{\gradk}[1]{\nabla f_{ik}(#1)}
\newcommand{\xk}{\mathbf{\theta}_{t}}
\newcommand{\xkk}{\mathbf{\theta}_{t+1}}
\newcommand{\fk}{\mathcal{L}_{\mathbb{B}_t}}
\newcommand{\normsq}[1]{\left\|#1\right\|^{2}}
\newcommand{\x}{\mathbf{\theta}}

\newcommand{\defequal}{:=}

\newcommandx{\BL}[1][1={,t}]{\mathcal{L}_{\mathbb{B}{#1}}}
\newcommandx{\DL}[1][1=t]{\mathcal{L}_{\mathbb{D}{#1}}}
\newcommandx{\Bl}[1][1={,t}]{l_{\mathbb{B}{#1}}}
\newcommandx{\Blh}[1][1={,t}]{\hat{l}_{\mathbb{B}{#1}}}
\newcommand{\argmin}[1]{\underset{#1}{\operatorname{arg}\,\operatorname{min}}\;}
\newcommand{\argmax}[1]{\underset{#1}{\operatorname{arg}\,\operatorname{max}}\;}
\renewcommandx{\L}{\mathcal{L}}
\newcommandx{\gb}[1][1={,t}]{\mathbf{g}_{\mathbb{B}#1}}
\newcommandx{\gbh}[1][1={,t}]{\hat{\mathbf{g}}_{\mathbb{B}#1}}
\newcommandx{\ngb}[1][1={,t}]{\hat{\mathbf{g}}_{\mathbb{B}#1}}

\newcommandx{\s}[1][1=\space]{s_{\text{upd}_{#1}}}
\newcommandx{\so}[1][1=\space]{s_{\text{opt}_{#1}}}
\renewcommandx{\mathbf}{\boldsymbol}
\newcommand{\pal}{\textit{PAL}}

\newcommand{\tikzremake}{\tikzset{external/force remake}}
\newcommand{\tikzdisable}{\tikzexternaldisable}
\newcommand{\tikzenable}{\tikzexternalenable}
	\title{Empirically explaining SGD from a line search perspective}
	\titlerunning{Empirically explaining SGD from a line search perspective (ICANN 2021)}
	\author{Maximus Mutschler\and
		Andreas Zell}
	%
	\institute{University of T\"ubingen, Sand 1, D-72076 T\"ubingen, Germany
		\\\email{\{maximus.mutschler, andreas.zell\}@uni-tuebingen.de}}
	\maketitle 
	\begin{abstract}
		Optimization in Deep Learning is mainly guided by vague intuitions and strong assumptions, with a limited understanding of how and why these work in practice. To shed more light on this, our work provides a deeper understanding of how SGD behaves by empirically analyzing the trajectory taken by SGD from a line search perspective. Specifically, a costly quantitative analysis of the full-batch loss along SGD trajectories from commonly used models trained on a subset of CIFAR-10 is performed. Our core results include that the full-batch loss along lines in update step direction is highly parabolically. Further on, we show that a learning rate exists with which SGD always performs almost exact line searches on the full-batch loss. Finally, we provide a new perspective on why increasing the batch size has almost the same effect as decreasing the learning rate by the same factor. 
	\end{abstract}
	\section{Introduction}
	Although the field of Deep Learning has made impressive progress in recent years, both in theory and application, little is known about why and how approaches work in detail. In general, Deep Learning approaches are based on vague intuitions in practice or rather strong assumptions in theory, \footnote{E.g., convexity, lipschitz continuity, interpolation, skip connections, batch normalization} without providing comprehensive empirical evidence that their intuitions and assumptions hold (e.g.: \cite{cyclicallearningrates,L4,backpropagation1,L4_alternative,sls,batchnorm,resnet,denseNet,vggnet}).\footnote{Better performance does not imply that the assumptions used are correct.} 
	Consequently, empirical analyses that search for a deeper understanding and explain why specific approaches work are rare to find.
	
	This is particularly valid for optimization, which, in this domain, is optimizing the mean of a stochastic loss function with an extremely high-dimensional parameter space. The landscape of such a loss function is generally assumed to be highly non-convex; however, recent works \cite{visualisationLossLandscape,walkwithsgd,pal,empericalLineSearchApproximations,probabilisticLineSearch,LinePlots,wedge_model,elasticband} claim that loss landscapes look rather simplistic for typical Deep Learning benchmarks used in optimization.\footnote{Image classification on MNIST, SVHN, CIFAR-10, CIFAR-100 and ImageNet} This is shown to be valid for the full-batch loss with low evidence and for mini-batch losses with stronger evidence. So far, there exists no detailed analysis of the relation of mini-batch losses to the full-batch loss to be optimized and of the actual performance of approaches using mini-batches on the full-batch loss. Globally, such an empirical analysis is not feasible in terms of resources and time, even if performed for a single model only. To nevertheless shed light on the subject, this work focuses on the quantitative analysis of full-batch and mini-batch losses along lines in SGD update step directions of a ResNet-20, a ResNet-18 \cite{resnet} and a MobileNetV2 \cite{mobilenet} trained on a computationally feasible subset of CIFAR-10 \cite{CIFAR-10}. Since the evaluation on each of the models supports our claims, we concentrate on the results of ResNet-20. \footnote{Results for the other models are given in the Appendix. We are aware that the analysis of a small set of problems provides low general evidence, but this is still better than no evidence at all. With the code published with this paper, it is simple to repeat our experiments on other problems.}
	
	Our core results are: 
	\textbf{1.} We provide further quantitative evidence that the full-batch loss along lines in update step direction behaves locally to a high degree parabolically (Sections \ref{la_sec_empirical_procedure},\ref{la_sec_shapeSimilarity}).
	\textbf{2.} We analyze the behavior of SGD \cite{backpropagation1}, parabolic approximation line search \cite{pal} and further approaches on the full-batch loss when trained on mini-batch losses (Section \ref{la_sec_lineLossApproximation}). We empirically show that there exists a leaning rate for which SGD always performs almost exact line searches on the full-batch loss. The former is since the optimal update step size on the full-batch loss and the norm of the gradient of the mini-batch loss behave approximately proportional.
	\textbf{3.} We consider the behavior of optimization approaches for different batch sizes (Section \ref{la_sec_influenceOfBatchSize}) and, from a new perspective, can quantitatively explain why increasing the batch size has virtually the same effect as decreasing the learning rate by the same factor, as experienced by \cite{decreaseLRincreaseBS}. 
	
	\section{Related work}
	\paragraph{\textbf{SGD trajectories:}}
	Similar to this work \cite{walkwithsgd} analyzes the loss along SGD trajectories, but with less focus on line searches and the exact shape of the full-batch loss. \cite{spectralnormalongsgdtrajectory} and \cite{hesssian_based_analysis_sgd} consider second-order information along SGD trajectories. Where \cite{spectralnormalongsgdtrajectory} investigates the spectral norm of the Hessian (highest curvature) along the SGD trajectory and shows, inter alia, that it initially visits increasingly sharp regions. \cite{hesssian_based_analysis_sgd} investigates the dynamics and generalization of SGD based on the Hessian of the loss. They show, among other things, that the primary subspace of the second momentum of stochastic gradients overlaps substantially with that of the Hessian. Thus, to an extent, SGD uses second-order information.
	
	\paragraph{\textbf{The simple loss landscape:}}
	Loss landscapes of Deep Learning problems can generally be highly non-convex, and thus, hard to optimize.
	In practice, however, loss landscapes tend to be simple: \cite{visualisationLossLandscape} suggests that loss landscapes of networks with skip connections behave smoothly. \cite{walkwithsgd} shows that the full-batch loss along SGD update step directions is roughly convex and that SGD bounces of walls of a \textit{valley like structure}. \cite{pal,empericalLineSearchApproximations} reveal that the batch loss along the update step direction is almost parabolically, and \cite{pal} suggests with weak empirical evidence that this also holds for the full-batch loss. Regarding this, \cite{probabilisticLineSearch} claims that the full-batch loss can be fitted by cubic splines along negative gradient directions. \cite{LinePlots} points out that on a straight path from initialization to solution, optimizers do not encounter any significant obstacles on the loss landscape. \cite{wedge_model} models the loss landscape as a set of high-dimensional wedges and demonstrates the existence of a low loss subspace connecting a set of minima. Similarly, \cite{elasticband} constructs continuous low-loss paths between minima and suggests that minima are best seen as points on single connected low-loss manifolds. 
	
	\paragraph{\textbf{Line searches:}} Recently, line searches have gained attention for optimization in Deep Learning. \cite{pal} shows empirically that a parabolic approximation line search on batch losses performs well across models and datasets. \cite{sls} proposes a simple, well-performing backtracking line search on mini-batch losses based on the interpolation assumption. The latter states that if the full-batch loss has zero gradient, then each mini-batch loss has zero gradient. \cite{probabilisticLineSearch} builds a local model of the full-batch loss along the update direction based on a Gaussian Process. 
	
	\paragraph{\textbf{Batch size and learning rate:}}
	Besides choosing the learning rate, selecting an appropriate batch size remains an important choice for SGD.
	\cite{ModelOfLargeBatchTraining} introduces the empirically-based "gradient noise scale", which predicts the largest beneficial batch size over datasets and models. \cite{bigBatchSGD} adaptively increases the batch size over update steps to assure that the negative gradient is a descent direction. \cite{decreaseLRincreaseBS} claims that decreasing the learning rate has virtually the same effect as increasing the batch size by the same factor. 
	
	\section{The empirical method}
	\label{la_sec_empirical_procedure}
	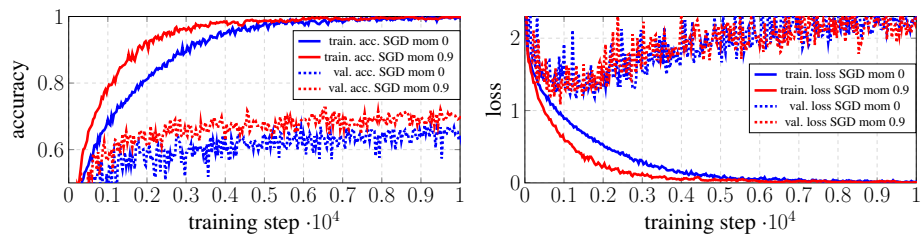
\begin{figure}[b]
	\newcommand\picscale{0.49}
	\newcommand\lossmin{0}
	\newcommand\lossmax{2.3}
	\newcommand\evalmin{0.5}
	\newcommand\evalmax{1}
	\newcommand\epochs{280}
	\newcommand\picheight{0.5\linewidth}
	\newcommand\legendwidth{9cm}
	\newcommand\legendheight{3cm}
\centering
		\tikzsetnextfilename{la_example_problem_accuracy}
		\begin{tikzpicture}[scale=\picscale] 
		\begin{axis}[
		width=\linewidth , 
		height=\picheight,
		grid=major      , 
		grid style={dashed,gray!30} , 
		xlabel=training step $\cdot 10^{4}$, 
		xlabel style={font=\LARGE},
		ylabel= accuracy,
		ylabel style={font=\LARGE},
		xmin=0,xmax=320,
		ymin=\evalmin,ymax=\evalmax ,
		scaled x ticks=true,
		      x coord trafo/.code={
		          \pgflibraryfpuifactive{
		              \pgfmathparse{(#1)*(31.25)*(0.0001)}
		          }{
		              \pgfkeys{/pgf/fpu=true}
		              \pgfmathparse{(#1)*(31.25)*(0.0001)}
		              \pgfkeys{/pgf/fpu=false}
		          }
		      },
		legend style={at={(0.573,0.5)},anchor=south west },
		x tick label style={rotate=0,anchor=near xticklabel,font=\LARGE}, 
		y tick label style={font=\LARGE},
		p1/.style={draw=blue,line width=2pt},
		p2/.style={draw=red,line width=2pt},
		p3/.style={draw=blue,line width=2pt,dotted},
		p4/.style={draw=red,line width=2pt,dotted},
		p5/.style={draw=green,line width=2pt,dotted},
		p6/.style={draw=green,line width=2pt},
		p7/.style={draw=black,line width=2pt,dotted},
		p8/.style={draw=black,line width=2pt},
		title style={font=\Huge}     ,
		]
		\addplot [p1] table[x=Step,y=Value,col sep=comma] {line_analysis/figure_data/figure1/train_acc00.csv};
		\addplot [p2] table[x=Step,y=Value,col sep=comma] {line_analysis/figure_data/figure1/train_acc09.csv}; 
		\addplot [p3] table[x=Step,y=Value,col sep=comma] 
		{line_analysis/figure_data/figure1/val_acc00.csv};
		\addplot [p4] table[x=Step,y=Value,col sep=comma] {line_analysis/figure_data/figure1/val_acc09.csv};

		\legend{train. acc. SGD mom 0, train. acc. SGD mom 0.9, val. acc. SGD mom 0, val. acc. SGD mom 0.9}
		\end{axis}
		\end{tikzpicture}
		\tikzsetnextfilename{la_example_problem_loss}
		\begin{tikzpicture}[scale=\picscale] 
		\begin{axis}[
		width=\linewidth, 
		height=\picheight,
		grid=major, 
		grid style={dashed,gray!30} , 
		xlabel= training step $\cdot 10^{4}$, 
		xlabel style={font=\LARGE},
		ylabel= loss,
		ylabel style={font=\LARGE} ,
		xmin=0,
		xmax=320,
		ymin=\lossmin,ymax=\lossmax,
      x coord trafo/.code={
          \pgflibraryfpuifactive{
              \pgfmathparse{(#1)*(31.25)*(0.0001)}
          }{
              \pgfkeys{/pgf/fpu=true}
              \pgfmathparse{(#1)*(31.25)*(0.0001)}
              \pgfkeys{/pgf/fpu=false}
          }
      },
		legend style={at={(0.573,0.3)},anchor=south west},
		x tick label style={rotate=0,anchor=near xticklabel,font=\LARGE}, 
		y tick label style={font=\LARGE},
		p1/.style={draw=blue,line width=2pt},
		p2/.style={draw=red,line width=2pt},
		p3/.style={draw=blue,line width=2pt,dotted},
		p4/.style={draw=red,line width=2pt,dotted},
		p5/.style={draw=green,line width=2pt,dotted},
		p6/.style={draw=green,line width=2pt},
		p7/.style={draw=black,line width=2pt,dotted},
		p8/.style={draw=black,line width=2pt},
		title style={font=\Huge}   ,
		]
		\addplot [p1] table[x=Step,y=Value,col sep=comma] {line_analysis/figure_data/figure1/train_loss00.csv};
		\addplot [p2] table[x=Step,y=Value,col sep=comma] {line_analysis/figure_data/figure1/train_loss09.csv}; 
		\addplot [p3] table[x=Step,y=Value,col sep=comma] 
		{line_analysis/figure_data/figure1/val_loss00.csv};
		\addplot [p4] table[x=Step,y=Value,col sep=comma] {line_analysis/figure_data/figure1/val_loss09.csv};

		\legend{train. loss SGD mom 0, train. loss SGD mom 0.9, val. loss SGD mom 0, val. loss SGD mom 0.9}
		\end{axis}
		\end{tikzpicture}

\caption{Training processes of a ResNet-20 trained on 8\% of CIFAR-10 with SGD with momentum 0 and 0.9. In the course of this work these processes will be analyzed in significant deeper details.}
\label{la_fig_training_process_resnet-20}
\end{figure}
	
\begin{figure}[b!]
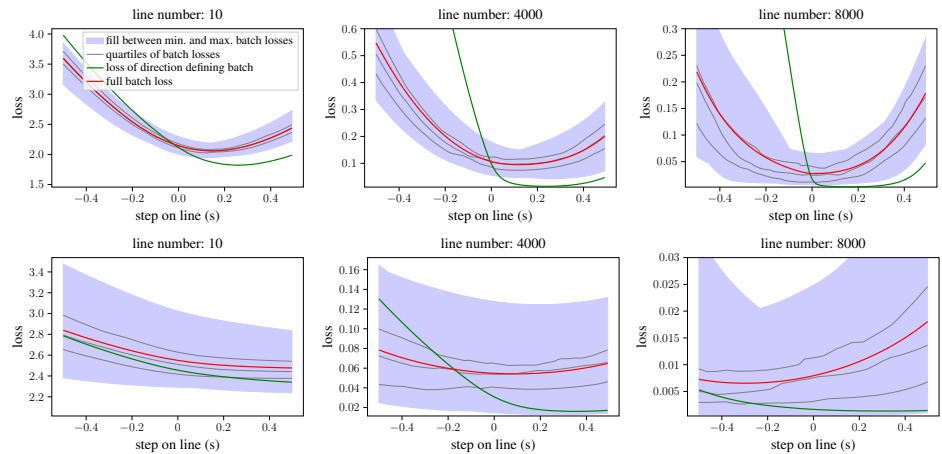

\centering
\def\scale{0.375}
\begin{tabular}{ c c c }
\tikzsetnextfilename{la_pure_line_10}
\scalebox{\scale}{\input{"line_analysis/figure_data/line_plots/CIFAR10_mom0_resnet20_augment_result/line_plots/bs_128_bs_ori_128_sgd_lrs_[0.01]_pal_mus_[0.1]_ri_100_c_apal_0.9_ori_sgd_lr_0.1/pure_line_10.pgf"}}&
\tikzsetnextfilename{la_pure_line_4000}
\scalebox{\scale}{\input{"line_analysis/figure_data/line_plots/CIFAR10_mom0_resnet20_augment_result/line_plots/bs_128_bs_ori_128_sgd_lrs_[0.01]_pal_mus_[0.1]_ri_100_c_apal_0.9_ori_sgd_lr_0.1/pure_line_4000.pgf"}}&
\tikzsetnextfilename{la_pure_line_8000}
\scalebox{\scale}{\input{"line_analysis/figure_data/line_plots/CIFAR10_mom0_resnet20_augment_result/line_plots/bs_128_bs_ori_128_sgd_lrs_[0.01]_pal_mus_[0.1]_ri_100_c_apal_0.9_ori_sgd_lr_0.1/pure_line_8000.pgf"}}\\
%
\tikzsetnextfilename{la_pure_line_10_mom}
\scalebox{\scale}{\input{"line_analysis/figure_data/line_plots/CIFAR10_mom09_resnet20_augment_result/line_plots/bs_128_bs_ori_128_sgd_lrs_[0.05]_pal_mus_[0.1]_ri_1008_c_apal_0.9_ori_sgd_lr_0.1/pure_line_10.pgf"}}&
\tikzsetnextfilename{la_pure_line_4000_mom}
\scalebox{\scale}{\input{"line_analysis/figure_data/line_plots/CIFAR10_mom09_resnet20_augment_result/line_plots/bs_128_bs_ori_128_sgd_lrs_[0.05]_pal_mus_[0.1]_ri_1008_c_apal_0.9_ori_sgd_lr_0.1/pure_line_4000.pgf"}}&\hspace{-0.1cm}
\tikzsetnextfilename{la_pure_line_8000_mom}
\scalebox{\scale}{\input{"line_analysis/figure_data/line_plots/CIFAR10_mom09_resnet20_augment_result/line_plots/bs_128_bs_ori_128_sgd_lrs_[0.05]_pal_mus_[0.1]_ri_1008_c_apal_0.9_ori_sgd_lr_0.1/pure_line_8000.pgf"}}
\end{tabular}

\caption{Losses along lines of the SGD training processes exhibit a simple shape. There is a significant difference between the full batch loss (red) and the loss of the direction defining batch (green). The loss of the direction defining batch is always steeper around 0. A mini-batch size of 128 is used. \textbf{Row 1:} SGD with momentum 0.0. \textbf{Row 2:} SGD with momentum 0.9. The mini-batch loss distributions exclude the direction defining mini batch.}
\label{la_fig_line_plots}
\end{figure}
 
	For the empirical analysis, a Deep Learning problem has to be chosen, which is (a) computationally so cheap that the analysis of the full-batch loss can be performed in a reasonable amount of time and (b) still is representative for typical Deep Learning benchmarks used in optimization.
	Therefore, this work considers the problem of training a ResNet-20\cite{resnet} on eight percent of the CIFAR-10 dataset\cite{CIFAR-10}. ResNet-like architectures are widely used in practice and CIFAR-10 is a commonly used baseline. The dataset is scaled down, so that computations for one training process take less than three weeks. Typical data augmentation is applied.\footnote{Cropping, horizontal flipping and normalization with mean and standard deviation.} Using PyTorch \cite{PyTorch}, the model is trained with SGD \cite{grad_descent} with learning rate $\lambda=0.1$,\footnote{Best performing $\lambda$ chosen of a grid search over $\{10^{-i}| i \in \{0,1,1.3,2,3,4\}\}$} batch size 128 and momentum $\beta$ of 0 and $0.9$ for 10000 steps.\
	
	Figure \ref{la_fig_training_process_resnet-20} shows the results of these SGD trainings. We note that the shown accuracies and losses do not provide much insight on what is happening on a deeper level. E.g. it does not provide much information why SGD performs well. To deal with this and further issues, the full-batch loss for each SGD update step is measured along lines in update step direction. This loss $l$ along direction $\mathbf{d}$ through the current parameters ${\theta}_0$ is given by: 
	\begin{equation}\label{la_eq_loss_along_line}
		l(s)=\mathcal{L}(\mathbf{\theta}_0+s\mathbf{d})=\frac{1}{|T|}\sum_{t \in T}L(t;\mathbf{\theta}_0+s\mathbf{d}),\end{equation}
	where $s$ is the step size along the line, $\mathcal{L}$ is the full-batch loss, $L$ is the sample loss and $T$ is the dataset. In the case of SGD without momentum, $\mathbf{d}$ is the negative unit gradient $-\mathbf{g}/||\mathbf{g}||$ of the original SGD trajectory whereas, in the case of SGD with momentum, $\mathbf{d}$ is the negative unit momentum direction $-\mathbf{m}/||\mathbf{m}||$. 
	
	For each of the 10000 update steps, we analyze the full-batch loss along the corresponding line in the interval $s\in[-0.5,0.5]$ 
	with a fine-grained resolution of 0.006. For each of the 167 sample step sizes along the line the sample loss of each element in the dataset is calculated. Then, all losses at a step size are averaged.
	All in all, this procedure requires more than 52 million inferences or 1.67 million epochs.
	
	Representative visualizations of mini- and full-batch losses along such lines are given in Figure \ref{la_fig_line_plots}. The following is observed considering all 10000 visualizations: The full-batch loss along lines has a simple, almost parabolic shape and does not change substantially across all lines. Further on is the slope of the direction defining mini-batch around $s=0$ is consistently steeper than the full-batch loss. The following sections provide further quantitative evidence that these observations hold. 
	
	In addition, we found the following interesting observations but do not investigate them further.
	There is a significant difference between the full-batch loss and the loss of the direction defining batch. Further, the loss of the direction defining batch does not follow the distribution of any other mini-batch loss along the line, especially for SGD without momentum. In addition, for SGD without momentum this loss is always lower and steeper than the other mini-batch losses. 
	\section{On the similarity of the shape of full-batch losses along lines}
	\label{la_sec_shapeSimilarity}

\begin{figure}[b!]
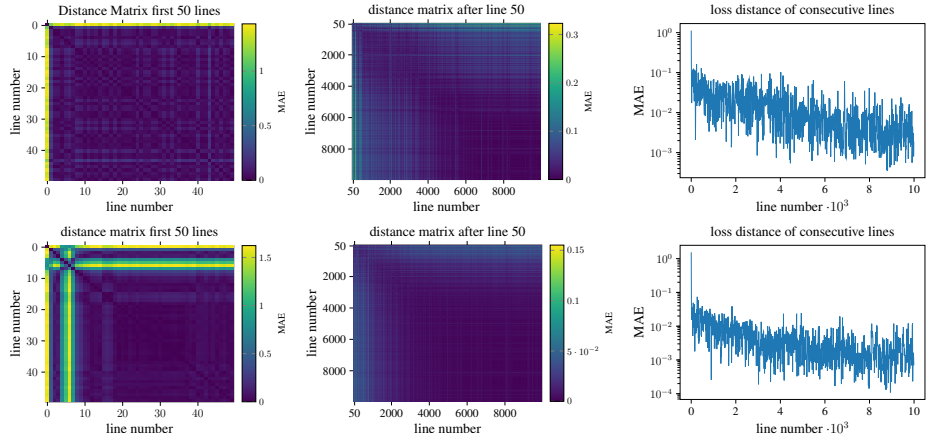

\centering
\def\scale{0.365}
\begin{tabular}{ c c c}

\tikzsetnextfilename{la_distance_matrix_first_50}
\scalebox{\scale}{
\begin{tikzpicture}

\begin{axis}[
yticklabel style={font=\large},
xticklabel style={font=\large},
ytick style={font=\Large},
xtick style={font=\Large},
ylabel style={font=\Large},
xlabel style={font=\Large},
colorbar,
colorbar style={ylabel={MAE}},
colormap/viridis,
point meta max=1.43428849352926,
point meta min=0,
tick align=outside,
title={Distance Matrix first 50 lines},
title style={font=\Large},
x grid style={white!69.0196078431373!black},
xlabel={line number},
xmin=-0.5, xmax=49.5,
xtick pos=both,
xtick style={color=black},
y dir=reverse,
y grid style={white!69.0196078431373!black},
ylabel={line number},
ymin=-0.5, ymax=49.5,
ytick pos=left,
ytick style={color=black}
]
\addplot graphics [includegraphics cmd=\pgfimage,xmin=-0.5, xmax=49.5, ymin=49.5, ymax=-0.5] {\figureTwoDataPath CIFAR10_mom0_resnet20_augment_result/statistics_plots/distance_matrix_first_50-000.png};
\end{axis}

\end{tikzpicture}}&
\tikzsetnextfilename{la_distance_matrix_after_line50}
\scalebox{\scale}{\hspace{-0.8cm}
\begin{tikzpicture}

\begin{axis}[
yticklabel style={font=\large},
xticklabel style={font=\large},
ytick style={font=\Large},
xtick style={font=\Large},
ylabel style={font=\Large},
xlabel style={font=\Large},
colorbar,
colorbar style={ylabel={MAE}},
colormap/viridis,
point meta max=0.323128379684872,
point meta min=0,
tick align=outside,
title={distance matrix after line 50 },
title style={font=\Large},
x grid style={white!69.0196078431373!black},
xlabel={line number},
xmin=-0.5, xmax=994.5,
xtick pos=both,
xtick style={color=black},
xtick={5,200,400,600,800},
xticklabels={50,2000,4000,6000,8000},
y dir=reverse,
y grid style={white!69.0196078431373!black},
ylabel={line number},
ymin=-0.5, ymax=994.5,
ytick pos=left,
ytick style={color=black},
ytick={5,200,400,600,800},
yticklabels={50,2000,4000,6000,8000}
]
\addplot graphics [includegraphics cmd=\pgfimage,xmin=-0.5, xmax=994.5, ymin=994.5, ymax=-0.5] {\figureTwoDataPath CIFAR10_mom0_resnet20_augment_result/statistics_plots/distance_matrix_after_line50-000.png};
\end{axis}

\end{tikzpicture}}&
\tikzsetnextfilename{la_consecutive_line_distances}
\scalebox{\scale}{\input{"line_analysis/figure_data/line_plots/CIFAR10_mom0_resnet20_augment_result/statistics_plots/consecutive_line_distances.pgf"}}\\

\tikzsetnextfilename{la_distance_matrix_first_50_mom}
\scalebox{\scale}{
\begin{tikzpicture}

\begin{axis}[
yticklabel style={font=\large},
xticklabel style={font=\large},
ytick style={font=\Large},
xtick style={font=\Large},
ylabel style={font=\Large},
xlabel style={font=\Large},
colorbar,
colorbar style={ylabel={MAE}},
colormap/viridis,
point meta max=1.62314299414926,
point meta min=0,
tick align=outside,
title={distance matrix first 50 lines},
title style={font=\Large},
x grid style={white!69.0196078431373!black},
xlabel={line number},
xmin=-0.5, xmax=49.5,
xtick pos=both,
xtick style={color=black},
y dir=reverse,
y grid style={white!69.0196078431373!black},
ylabel={line number},
ymin=-0.5, ymax=49.5,
ytick pos=left,
ytick style={color=black}
]
\addplot graphics [includegraphics cmd=\pgfimage,xmin=-0.5, xmax=49.5, ymin=49.5, ymax=-0.5] {\figureTwoDataPath CIFAR10_mom09_resnet20_augment_result/statistics_plots/distance_matrix_first_50-000.png};
\end{axis}

\end{tikzpicture}}&
\tikzsetnextfilename{la_distance_matrix_after_line50_mom}
\scalebox{\scale}{
\begin{tikzpicture}

\begin{axis}[
yticklabel style={font=\large},
xticklabel style={font=\large},
ytick style={font=\Large},
xtick style={font=\Large},
ylabel style={font=\Large},
xlabel style={font=\Large},
colorbar,
colorbar style={ylabel={MAE}},
colormap/viridis,
point meta max=0.155194373616511,
point meta min=0,
tick align=outside,
title={distance matrix after line 50 },
title style={font=\Large},
x grid style={white!69.0196078431373!black},
xlabel={line number},
xmin=-0.5, xmax=994.5,
xtick pos=both,
xtick style={color=black},
xtick={5,200,400,600,800},
xticklabels={50,2000,4000,6000,8000},
y dir=reverse,
y grid style={white!69.0196078431373!black},
ylabel={line number},
ymin=-0.5, ymax=994.5,
ytick pos=left,
ytick style={color=black},
ytick={5,200,400,600,800},
yticklabels={50,2000,4000,6000,8000}
]
\addplot graphics [includegraphics cmd=\pgfimage,xmin=-0.5, xmax=994.5, ymin=994.5, ymax=-0.5] {\figureTwoDataPath CIFAR10_mom09_resnet20_augment_result/statistics_plots/distance_matrix_after_line50-000.png};
\end{axis}

\end{tikzpicture}}& 
\tikzsetnextfilename{la_consecutive_line_distances_mom}
\scalebox{\scale}{\input{"line_analysis/figure_data/line_plots/CIFAR10_mom09_resnet20_augment_result/statistics_plots/consecutive_line_distances.pgf"}}
\end{tabular}
\caption{Distances of the shape of full-batch losses along lines in a window around the current position $s=0$.
	\textbf{Row 1:} SGD without momentum. \textbf{Row 2:} SGD with momentum. Since the offset is not of interest the minimum is shifted to 0 on the y-axis. The distances are rather high for the first 10 lines (left). For the following lines the distances are less than 0.3 MAE (middle) and concentrate around 0.01. The MAEs of the full-batch loss of pairs of consecutive lines are given on the right.}
\label{la_fig_similarity_matrices}
\end{figure}
\begin{figure}[t!]
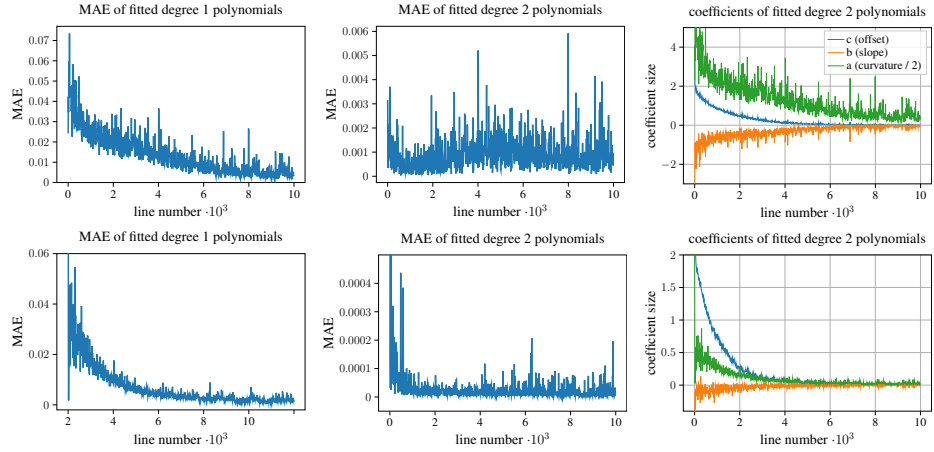

\centering
\def\scale{0.37}
\begin{tabular}{ c c c}
\tikzsetnextfilename{la_mae_of_polynomial_fit_of_degree_1}
\scalebox{\scale}{\input{"line_analysis/figure_data/line_plots/CIFAR10_mom0_resnet20_augment_result/statistics_plots/mae_of_polynomial_fit_of_degree_1.pgf"}}&
\tikzsetnextfilename{la_mae_of_polynomial_fit_of_degree_2}
\scalebox{\scale}{\input{"line_analysis/figure_data/line_plots/CIFAR10_mom0_resnet20_augment_result/statistics_plots/mae_of_polynomial_fit_of_degree_2.pgf"}}&
\tikzsetnextfilename{la_coefficients_of_polynomial_fit_of_degree_2}
\scalebox{\scale}{\input{"line_analysis/figure_data/line_plots/CIFAR10_mom0_resnet20_augment_result/statistics_plots/coefficients_of_polynomial_fit_of_degree_2.pgf"}}\\
\tikzsetnextfilename{la_mae_of_polynomial_fit_of_degree_1_mom}
	\scalebox{\scale}{\input{"line_analysis/figure_data/line_plots/CIFAR10_mom09_resnet20_augment_result/statistics_plots/mae_of_polynomial_fit_of_degree_1.pgf"}}&
\tikzsetnextfilename{la_mae_of_polynomial_fit_of_degree_2_mom}
	\scalebox{\scale}{\input{"line_analysis/figure_data/line_plots/CIFAR10_mom09_resnet20_augment_result/statistics_plots/mae_of_polynomial_fit_of_degree_2.pgf"}}&
\tikzsetnextfilename{la_coefficients_of_polynomial_fit_of_degree_2_mom}
	\scalebox{\scale}{\input{"line_analysis/figure_data/line_plots/CIFAR10_mom09_resnet20_augment_result/statistics_plots/coefficients_of_polynomial_fit_of_degree_2.pgf"}}
\end{tabular}
\caption{MAE of polynomial approximations of the full-batch loss of degree 1 and 2. \textbf{Row 1:} SGD without momentum. \textbf{Row 2:} SGD with momentum. Full-batch losses along lines can be well fitted by polynomials of degree 2. The slope of the approximation stays roughly constant whereas the curvature decreases.}
\label{la_fig_polynomial_approximations}
\end{figure}

	The visualization of the full-batch loss along 10000 lines suggests that the shape of this loss does not vary significantly during the training process. For a more detailed investigation, the Mean Absolute Error (MAE) of the full-batch loss between each pair of lines is analyzed on a relevant interval. Since solely the shape of the loss is of interest and not the offset, each loss along a line is shifted along the y-axis, such that the minimum is at zero. The interval from $s\in[-0.2,0.2]$ is considered for SGD and from $s\in[-0.5,0.5]$ for SGD with momentum. The latter ensures that the minimum position and the origin are always included. The resulting distance matrices are depicted in Figure \ref{la_fig_similarity_matrices}. They show that \textbf{only the shapes of the full-batch loss of the very first lines vary strongly, whereas, later shapes behave more alike. In particular, the full-batch loss along consecutive lines behaves similarly.} This favors optimization with fixed step sizes, since the optimal update step does not change much. These results are also valid for the full-batch loss along each line in multiple noisy gradient directions starting from the same position in parameter space (Appendix Figure \ref{la_fig_similarity_matrices_on_position}). This implies from an optimization point of view that it does not matter which of the descent directions is taken.\\
	\indent Figure \ref{la_fig_line_plots} also indicates that \textbf{the full-batch loss along lines exhibits an almost parabolic shape locally} (core result 1). Figure \ref{la_fig_polynomial_approximations} shows in detail that this is valid since the fitting error of a parabola is always low. In addition, we can see that the curvature of the fitted parabolas (i.e., the second directional derivative) decreases during training. This implies that \textbf{the approximated loss becomes flatter and suggests that SGD follows a simple valley-like structure which becomes continuously wider}. Considering the even faster-decreasing curvature of SGD with momentum, its valley becomes even wider (see also Figure \ref{la_fig_line_plots}). This might be a reason why SGD with momentum optimizes and generalizes better \cite{large_batch_wide_minima,hochreiter_wide_minima}. In accordance with \cite{spectralnormalongsgdtrajectory}, we also found that the curvature is increasing rapidly during the very first steps and then decreases. \\
	\indent Supporting results are obtained for ResNet-18 and for MobileNetV2 see Appendix Figures \ref{la_fig_distance_matrix_resnet18}, \ref{la_fig_distance_matrix_resnet18_2}, \ref{la_fig_resnet_18_polynomial_approximations}, \ref{la_fig_mobilenet_polynomial_approximations}.
	
	\FloatBarrier
	\section{On the behavior of line search approaches on the full-batch loss}
	\label{la_sec_lineLossApproximation}
	The previous section showed that the full-batch loss along lines in update step direction behaves parabolically and exhibits positive curvature. This means that $l(s)\approx as^2+bs+c$ with $a>0$ (see Equation \ref{la_eq_loss_along_line}). In the following, the performance of several parabolic approximation line searches applied on the direction defining mini-batch loss are analyzed. From now on, we concentrate on SGD without momentum, but, Figure \ref{la_fig_strategy_metrics_momentum} (Appendix) shows that the upcoming results for SGD with momentum mostly support the derivations.
	
	For SGD the mini-batch loss and its gradient $\mathbf{g}$ are given at the origin ($s=0$) of a line. In addition, the directional derivative, which is the negative norm of $\mathbf{g}$, can be computed easily ($\text{-} \nicefrac{\mathbf{g}}{||\mathbf{g}||}\cdot \mathbf{g}^T= \text{-}||\mathbf{g}||$). To perform a parabolic approximation, either one additional loss along the line has to be considered or the curvature has to be estimated. The first approach is proposed by \cite{pal}. The default update step of their optimizer PAL is given as: \begin{figure}[b!]
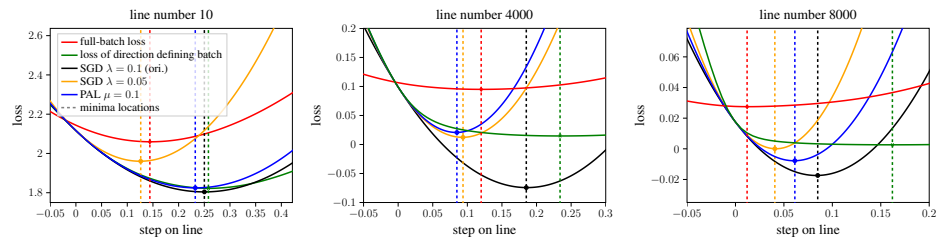

\def\scale{0.361}
\begin{tabular}{ c c c}
	\tikzsetnextfilename{la_sgd_pal_line_10}
	\scalebox{\scale}{\input{"line_analysis/figure_data/line_plots/CIFAR10_mom0_resnet20_augment_result/line_plots/bs_128_bs_ori_128_sgd_lrs_[0.05]_pal_mus_[0.1]_ri_50_c_apal_0.9_ori_sgd_lr_0.1/sgd_pal_line_10.pgf"}}&
	\tikzsetnextfilename{la_sgd_pal_line_4000}
	\scalebox{\scale}{\input{"line_analysis/figure_data/line_plots/CIFAR10_mom0_resnet20_augment_result/line_plots/bs_128_bs_ori_128_sgd_lrs_[0.05]_pal_mus_[0.1]_ri_50_c_apal_0.9_ori_sgd_lr_0.1/sgd_pal_line_4000.pgf"}}&
	\tikzsetnextfilename{la_sgd_pal_line_8000}
	\scalebox{\scale}{\input{"line_analysis/figure_data/line_plots/CIFAR10_mom0_resnet20_augment_result/line_plots/bs_128_bs_ori_128_sgd_lrs_[0.05]_pal_mus_[0.1]_ri_50_c_apal_0.9_ori_sgd_lr_0.1/sgd_pal_line_8000.pgf"}}
\end{tabular}


\caption{Several parabolic line approximations and their minimum positions on representative losses along lines.  The optimal update step, from a local perspective, is depicted by the red dashed line. The other update steps are derived from the direction defining mini-batch loss.}

\label{la_fig_line_approximations}
\end{figure}
	\begin{figure}[t!]
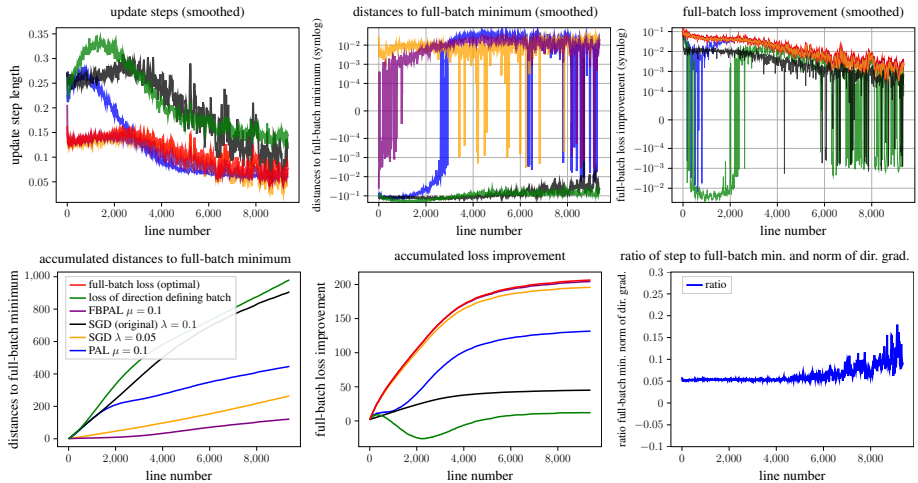

	\centering
	\def\scale{0.362}
	\def\basis{10e-6}

	\begin{tabular}{ c c c }
		\tikzsetnextfilename{la_strategy_metrics_update_steps}
		\scalebox{\scale}{\input{"line_analysis/figure_data/line_plots/CIFAR10_mom0_resnet20_augment_result/statistics_plots/bs_128_bs_ori_128_sgd_lrs_[0.05]_pal_mus_[0.1]_ri_100_c_apal_0.9_ori_sgd_lr_0.1/step_step.pgf"}}&\hspace{-0.25cm}
		\pgfplotsset
		{
			y coord trafo/.code={\pgfmathparseFPU{symlog(#1,\basis)}},
			y coord inv trafo/.code={\pgfmathparseFPU{symexp(#1,\basis)}},
		}
		\tikzsetnextfilename{la_strategy_metrics_distances}
		\scalebox{\scale}{\input{"line_analysis/figure_data/line_plots/CIFAR10_mom0_resnet20_augment_result/statistics_plots/bs_128_bs_ori_128_sgd_lrs_[0.05]_pal_mus_[0.1]_ri_100_c_apal_0.9_ori_sgd_lr_0.1/distances_step.pgf"}}&\hspace{-0.25cm}
		\pgfplotsset
		{
		y coord trafo/.code={\pgfmathparseFPU{symlog(#1,\basis)}},
		y coord inv trafo/.code={\pgfmathparseFPU{symexp(#1,\basis)}},
		}\tikzsetnextfilename{la_strategy_metrics_improvements}
		\scalebox{\scale}{\input{"line_analysis/figure_data/line_plots/CIFAR10_mom0_resnet20_augment_result/statistics_plots/bs_128_bs_ori_128_sgd_lrs_[0.05]_pal_mus_[0.1]_ri_100_c_apal_0.9_ori_sgd_lr_0.1/improvements_step.pgf"}}\\
		\tikzsetnextfilename{la_strategy_metrics_distance_accumulated}
		\scalebox{\scale}{\input{"line_analysis/figure_data/line_plots/CIFAR10_mom0_resnet20_augment_result/statistics_plots/bs_128_bs_ori_128_sgd_lrs_[0.05]_pal_mus_[0.1]_ri_100_c_apal_0.9_ori_sgd_lr_0.1/distances_step_accumulated.pgf"}}&\hspace{-0.25cm}
		\tikzsetnextfilename{la_strategy_metrics_improvement_accumulated}
		\scalebox{\scale}{\input{"line_analysis/figure_data/line_plots/CIFAR10_mom0_resnet20_augment_result/statistics_plots/bs_128_bs_ori_128_sgd_lrs_[0.05]_pal_mus_[0.1]_ri_100_c_apal_0.9_ori_sgd_lr_0.1/improvements_step_accumulated.pgf"}}&\hspace{-0.25cm}
		\tikzsetnextfilename{la_strategy_metrics_ratio_fb_min_mb_grad}
		\scalebox{\scale}{\input{"line_analysis/figure_data/line_plots/CIFAR10_mom0_resnet20_augment_result/statistics_plots/bs_128_bs_ori_128_sgd_lrs_[0.05]_pal_mus_[0.1]_ri_100_c_apal_0.9_ori_sgd_lr_0.1/fb_min_step_ratio.pgf"}}
	\end{tabular}

	\caption{Several metrics to compare update step strategies: 1. update step sizes. 2. the distance to the minimum of the full batch loss ($s_{opt}-s_{upd}$), which is the optimal update step from a local perspective. 3. the loss improvement per step given as:  $l(0)-l(s_{upd})$ where $s_{upd}$ is the update step of a strategy. Average smoothing with a kernel size of 25 is applied. The right lower plot shows almost proportional behavior between $s_{opt}$ and the directional derivative of the direction defining mini-batch loss. }
	\label{la_fig_strategy_metrics}
\end{figure}
	\begin{equation}\label{la_eq_update_step} 
		s_{\text{pal}}=-\frac{b}{2a}=-\frac{\Bl[]^{\prime}(0)\mu^2}{2(\Bl[](\mu)-\Bl[](0)-\Bl[]^{\prime}(0)\mu)}\phantom{a},
	\end{equation}where $\Bl[]$ is the mini-batch loss along a line in the direction of $\mathbf{g}$ and $\mu$ is the sample step size for the second loss.
	The second approach is a reinterpretation of SGD as a parabolic approximation line search with estimated curvature. SGD's update step is given as $-\lambda \mathbf{g}$, where $\lambda$ is the learning rate. Considering a normalized gradient and defining $k=\frac{1}{\lambda}$ as the curvature, we get
	\begin{equation}\label{la_eq_sgd_interpretation}
		\footnotesize
		-\lambda \mathbf{g}=\lambda ||\mathbf{g}||\cdot \frac{-\mathbf{g}}{||\mathbf{g}||}
		=\frac{||\mathbf{g}||}{k}\cdot \frac{-\mathbf{g}}{||\mathbf{g}||}
		=-\frac{\frac{-\mathbf{g}}{||\mathbf{g}||}\mathbf{g}^T}{k}\cdot \frac{-\mathbf{g}}{||\mathbf{g}||}
		=- \frac{\text{\scriptsize{first directional derivative}}}{\text{\scriptsize{second directional derivative}}} \cdot\text{\scriptsize{direction}}\end{equation}
	Note that the latter is a Newton update step.
	
	To get a first intuition of how these approaches operate, several parabolic approximations and their resulting update steps on representative lines are shown in Figure\ref{la_fig_line_approximations}.
	
	The next step is to compare several update step strategies using three metrics. Beforehand, we have to define $\so$ as the step size to the minimum of the full-batch loss along a line, which is the optimal update step size from a local perspective. $s_{upd}$ is the update step size of an arbitrary optimization strategy considered. The metrics are: the update step size $s_{upd}$, the distance of $s_{upd}$ to the minimum of the full-batch loss ($\so-s_{upd}$), and the loss improvement per step, given as: $l(0)-l(s_{upd})$, where $l$ is the full-batch loss along a line (see Equation \ref{la_eq_loss_along_line}). Note that this improvement measure does not represent actual training performance since the next considered line is independent of the previous update step size for all strategies except for SGD, which training process we are considering. However, it does represent the performance on full-batch losses along lines, which are likely to occur during training.\\
	Figure \ref{la_fig_strategy_metrics} shows that some strategies exhibit varying behavior on the metrics. To strengthen our previous observation, a parabolic approximation on the full-batch loss (FBPAL) yields almost optimal performance. Surprisingly, SGD with $\lambda=0.05$ estimates the minima of the full-batch loss almost as well. This is because \textbf{the step to the minimum of the full-batch loss $\so$ is almost proportional to the directional derivative ($-||\mathbf{g}||$) of the direction defining mini-batch loss} (core result 2), as shown in the lower plot of Figure \ref{la_fig_strategy_metrics}. Observe that the variance becomes larger during the end of the training, and thus the proportionality holds less. \textbf{This almost proportional behavior explains why a constant learning rate can lead to a good performance, since it is sufficient to control the update step size with the norm of the noisy mini-batch gradient}. In practice, however, this locally optimal learning rate is unknown. The globally best performing learning rate of $0.1$ always does a step far beyond the locally optimal step. The latter is what \cite{walkwithsgd} described as \textit{bouncing off walls of a valley-like structure}. Contrary to their intuition, we have not found any boundaries at all in the valley. Finally, Figure \ref{la_fig_strategy_metrics} suggests that \textbf{exact line searches on the mini-batch loss perform poorly}.\\
	Supporting results are obtained for SGD with momentum, for ResNet-18 and for MobileNetV2 see Appendix Figures \ref{la_fig_strategy_metrics_momentum}, \ref{la_fig_strategy_metrics_res_net18}, \ref{la_fig_strategy_metrics_res_net18_2}, \ref{la_fig_strategy_metrics_res_net18_3}, \ref{la_fig_strategy_metrics_res_net18_4}. However, in the case of SGD with momentum the line search is constantly not as exact.\\
	Combining the last core results suggest that \textbf{the locally optimal step size} $\so$ \textbf{can be well approximated by a Newton step on the full-batch loss or by a simple proportionality}:\begin{equation}
		\so \approx -\frac{-||\mathbf{g}_{fbl}||}{\mathbf{g}_{fbl}H_{fbl}\mathbf{g}_{fbl}^T}\approx c\cdot -||\mathbf{g}_{dl}||\end{equation}
	where $fbl$ stands for the full-batch loss and $dl$ for the loss of the direction defining mini-batch. However, \textbf{on a global perspective a step size larger than} $\so$\textbf{, can perform better, although it yields locally lower improvement} (Appendix Figure \ref{la_fig_sgd_overshoot} ).
	
	\FloatBarrier
	\section{On the influence of the batch size on update steps}
	\label{la_sec_influenceOfBatchSize}
	\begin{figure}[b!]
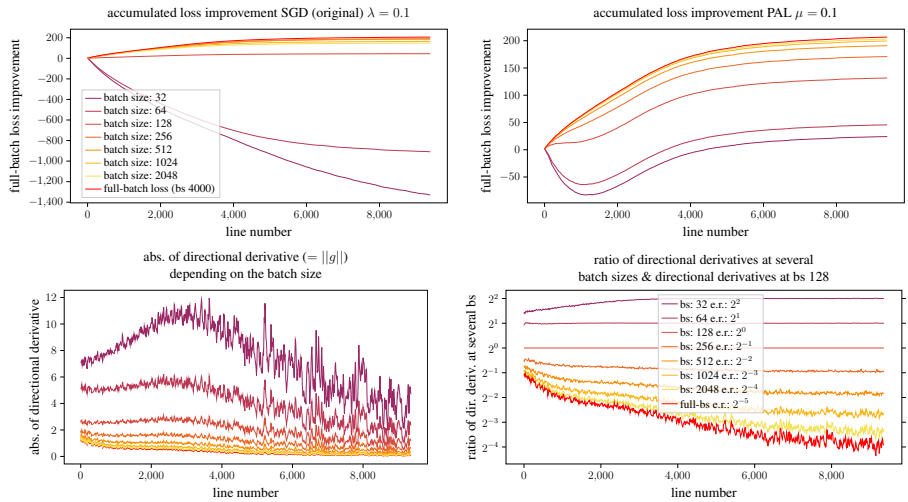

	\centering
	\def\scale{0.36}
	\begin{tabular}{ c c}
	\tikzsetnextfilename{la_batch_size_comparison_sgd_improvement}
	\scalebox{\scale}{\input{"line_analysis/figure_data/line_plots//CIFAR10_mom0_resnet20_augment_result/statistics_plots/batch_size_comparison/SGDoriginallambda=0,1_improvements_step_accumulated.pgf"}}&\hspace{0.1cm}	
	\tikzsetnextfilename{la_batch_size_comparison_pal_improvement}
	\scalebox{\scale}{\input{"line_analysis/figure_data/line_plots/CIFAR10_mom0_resnet20_augment_result/statistics_plots/batch_size_comparison/PALmu=0,1_improvements_step_accumulated.pgf"}}\\
	\tikzsetnextfilename{la_batch_size_comparison_directional_derivatives}
	\scalebox{\scale}{\input{"line_analysis/figure_data/line_plots/CIFAR10_mom0_resnet20_augment_result/statistics_plots/batch_size_comparison/directional_derivatives.pgf"}}&
	\tikzsetnextfilename{la_batch_size_comparison_directional_derivatives_ratio}
	\scalebox{\scale}{\input{"line_analysis/figure_data/line_plots/CIFAR10_mom0_resnet20_augment_result/statistics_plots/batch_size_comparison/directional_derivatives_ratio.pgf"}}
	\end{tabular}\\
	\caption{\textbf{Row 1} Comparing the influence of the batch size on the loss improvement. Left: SGD with the original learning rate of 0.1. Right: parabolic approximation line search (PAL).
	\textbf{Row 2: } Analysis of the relation of the batch size to the absolute directional derivative (=gradient norm) which shows in detail that increasing the batch size has a similar effect as decreasing the learning rate by the same factor.  \textbf{e.r.}  stands for expected ratio.}.
	\label{la_fig_distances_batchwise}
\end{figure}
 
	This section analyzes to which extent the performance of SGD and PAL changes with varying batch sizes. In addition, we show why, on the losses along lines measured, increasing the batch size has almost the same effect as decreasing the learning rate by the same factor, as suggested by \cite{decreaseLRincreaseBS}.
	
	The presented results are simplified, assuming that the SGD trajectory keeps identical with changing batch size. Thus, the same losses over lines can be considered. The original batch size is 128. For larger batch sizes, additional sample losses from the set of all measured losses are drawn without replacement. For smaller batch sizes, the sample losses with the highest directional derivatives are removed, assuming that for smaller batch sizes steeper steepest directions are found.
	
	The upper plots of Figure \ref{la_fig_distances_batchwise} show that SGD performs significantly worse for smaller batch sizes than PAL does. Both approaches become significantly more accurate at larger batch sizes. A batch size of 512 is already sufficient to perform almost optimally.
	
	\cite{decreaseLRincreaseBS} shows that when training a ResNet-50\cite{resnet} on ImageNet \cite{IMAGENET}, increasing the batch size has virtually the same effect as decreasing the learning rate by the same factor. Their interpretation is based on the noise on the full-batch gradient introduced by mini-batches whereas, we argue from the perspective of mini-batch losses. The SGD update step length on losses along a line is the absolute of the learning rate times the directional derivative ($\lambda \cdot |l'_m(0)|=\lambda \cdot ||\mathbf{g}||$). The lower left plot of Figure \ref{la_fig_distances_batchwise} shows that with larger batch sizes, the absolute of the directional derivative, and thus the step size, decreases. This can be figuratively explained with the help of Figure \ref{la_fig_line_plots}. As the batch size increases, the loss of the direction defining batch becomes more similar to the full-batch loss; consequently, the absolute of the directional derivative decreases. The lower plot of Figure \ref{la_fig_distances_batchwise} shows by which factor the directional derivative is divided when the batch size is multiplied by a factor. \textbf{For batch size 32 to 256 the assumption that if the batch size is increased by a factor, then the update step size decreases by the same factor, is valid during the whole training} (core result 3). For larger batch sizes, the directional derivative is divided by a lower factor at the beginning of the training, then the batch size is multiplied but converges towards the same factor during the training. Based on the data collected, we cannot estimate the momentum term for a different batch size for each line; therefore, this analysis was not performed for SGD with momentum. Supporting results are obtained on ResNet-18 \cite{resnet} and a MobileNetV2 \cite{mobilenet} see appendix Section \ref{la_app_resnet18_mobilenet} Figure \ref{la_fig_distances_batchwise_app} and \ref{la_fig_distances_batchwise_app2}.
	\vspace{-0.2cm}
	\section{Discussion and Outlook}
	\vspace{-0.2cm}
	With this work, we provided a better understanding of what happens in detail during SGD training from a line search perspective. In short, we quantitatively showed that the full-batch loss along lines in update step direction locally is highly parabolically. Further on, we found a learning rate for which SGD always performs an almost optimal line search. This questions whether line searches for deep learning can ever outperform SGD in general. Finally, we quantitatively analyzed the relation of learning rate and batch size in detail and provided a new perspective on why increasing the batch size has almost the same effect as decreasing the learning rate by the same factor.

	We have to emphasize that this work focused on a small set of representative problems only. Therefore, our results have to be handled with care. To get a more general view about the behavior of SGD and other optimizers across models and datasets, we propose to repeat these or similar experiments for as many as possible. This can be easily done with the published code but is extraordinarily time-consuming (see {\scriptsize \url{https://github.com/cogsys-tuebingen/empirically_explaining_sgd_from_a_line_search_perspective}}). 
	
	In general, we want to emphasize that a prospective goal of future studies in Deep Learning should be, beyond reporting good results, to provide empirical evidence that the assumptions used hold.

	{
		\bibliographystyle{splncs04}
		\bibliography{line_analysis}
	}
	\vfill
	\pagebreak
	\onecolumn
	\appendix
	\section{Further results on ResNet-20}
	\label{la_app_res_net_20} 
	\FloatBarrier
	\begin{figure}[h!]
	\centering
	\def\scale{0.38}
	\begin{tabular}{ c c c}	
		\tikzsetnextfilename{la_res_net_20_distance_matrix_on_pos_0}
		\scalebox{\scale}{
\begin{tikzpicture}

\begin{axis}[
colorbar,
colorbar style={ylabel={MAE}},
colormap/viridis,
point meta max=0.279651901590537,
point meta min=0,
tick align=outside,
title={Distance Matrix line 0},
title style={font=\Large},
x grid style={white!69.0196078431373!black},
xlabel={line number},
xmin=-0.5, xmax=30.5,
xtick pos=both,
xtick style={color=black},
y dir=reverse,
y grid style={white!69.0196078431373!black},
ylabel={line number},
ymin=-0.5, ymax=30.5,
ytick pos=left,
ytick style={color=black},
yticklabel style={font=\large},
xticklabel style={font=\large},
ytick style={font=\Large},
ylabel style={font=\Large},
xlabel style={font=\Large},
]
\addplot graphics [includegraphics cmd=\pgfimage,xmin=-0.5, xmax=30.5, ymin=30.5, ymax=-0.5] {\figureTwoDataPath CIFAR10_mom0_resnet20_augment_on_pos_result/statistics_plots/group_0/distance_matrix-000.png};
\end{axis}

\end{tikzpicture}}&
	\tikzsetnextfilename{la_res_net_20_distance_matrix_on_pos_4000}
	\scalebox{\scale}{
\begin{tikzpicture}

\begin{axis}[
colorbar,
colorbar style={ylabel={MAE}},
colormap/viridis,
point meta max=0.0478017910052716,
point meta min=0,
tick align=outside,
title={Distance Matrix line 4000},
title style={font=\Large},
x grid style={white!69.0196078431373!black},
xlabel={line number},
xmin=-0.5, xmax=30.5,
xtick pos=both,
xtick style={color=black},
y dir=reverse,
y grid style={white!69.0196078431373!black},
ylabel={line number},
ymin=-0.5, ymax=30.5,
ytick pos=left,
ytick style={color=black},
yticklabel style={font=\large},
xticklabel style={font=\large},
ytick style={font=\Large},
ylabel style={font=\Large},
xlabel style={font=\Large},
]
\addplot graphics [includegraphics cmd=\pgfimage,xmin=-0.5, xmax=30.5, ymin=30.5, ymax=-0.5] {\figureTwoDataPath CIFAR10_mom0_resnet20_augment_on_pos_result/statistics_plots/group_4000/distance_matrix-000.png};
\end{axis}

\end{tikzpicture}}&
	
	\tikzsetnextfilename{la_res_net_20_distance_matrix_on_pos_8000}
	\scalebox{\scale}{
\begin{tikzpicture}

\begin{axis}[
colorbar,
colorbar style={ylabel={MAE}},
colormap/viridis,
point meta max=0.00775389332038998,
point meta min=0,
tick align=outside,
title={Distance Matrix line 8000},
title style={font=\Large},
x grid style={white!69.0196078431373!black},
xlabel={line number},
xmin=-0.5, xmax=30.5,
xtick pos=both,
xtick style={color=black},
y dir=reverse,
y grid style={white!69.0196078431373!black},
ylabel={line number},
ymin=-0.5, ymax=30.5,
ytick pos=left,
yticklabel style={font=\large},
xticklabel style={font=\large},
ytick style={font=\Large},
ylabel style={font=\Large},
xlabel style={font=\Large},
]
\addplot graphics [includegraphics cmd=\pgfimage,xmin=-0.5, xmax=30.5, ymin=30.5, ymax=-0.5] {\figureTwoDataPath CIFAR10_mom0_resnet20_augment_on_pos_result/statistics_plots/group_8000/distance_matrix-000.png};
\end{axis}

\end{tikzpicture}}
	\end{tabular}
	\vspace{-0.4cm}
	\caption{Distances (MAE) of the shape of full-batch losses along lines in multiple noisy gradient direction in a window of 0.3 around the line origin $s=0$. The minimum is shifted to 0 on the y-axis. At fixed positions in parameter space the full-batch loss along lines in several noisy gradient directions reveals low distances. Those plots are representative for 100 positions we analyzed.}
	\label{la_fig_similarity_matrices_on_position}
\end{figure}
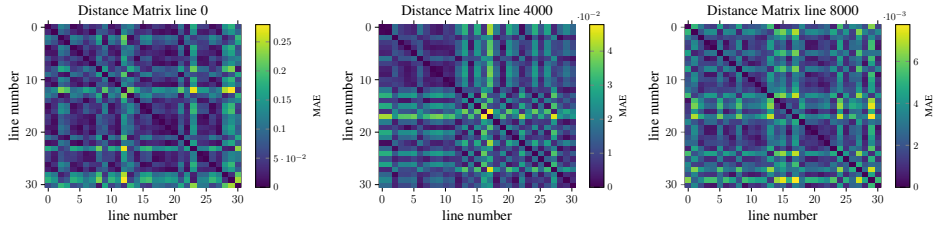
	\begin{figure}[h!]
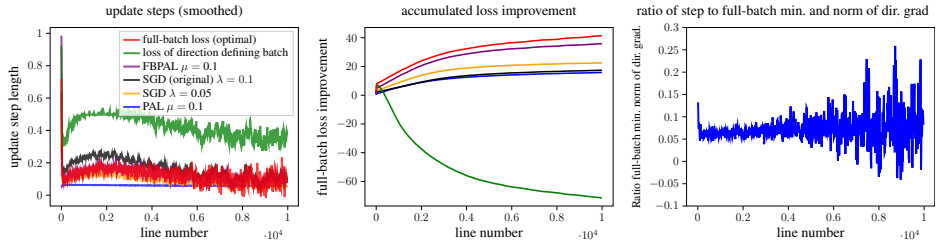

	\centering

	\def\scale{0.37}
	\def\basis{10e-5}
	\begin{tabular}{ c c c }
		\tikzsetnextfilename{la_strategy_metrics_update_steps_mom}
		\scalebox{\scale}{\input{"line_analysis/figure_data/line_plots/CIFAR10_mom09_resnet20_augment_result/statistics_plots/bs_128_bs_ori_128_sgd_lrs_[0.05]_pal_mus_[0.1]_ri_100_c_apal_0.9_ori_sgd_lr_0.1/step_step.pgf"}}&
	
		\tikzsetnextfilename{la_strategy_metrics_improvement_accumulated_mom}
		\scalebox{\scale}{\input{"line_analysis/figure_data/line_plots/CIFAR10_mom09_resnet20_augment_result/statistics_plots/bs_128_bs_ori_128_sgd_lrs_[0.05]_pal_mus_[0.1]_ri_100_c_apal_0.9_ori_sgd_lr_0.1/improvements_step_accumulated.pgf"}}&
		\tikzsetnextfilename{la_strategy_metrics_ratio_fb_min_mb_grad_mom}
		\scalebox{\scale}{\input{"line_analysis/figure_data/line_plots/CIFAR10_mom09_resnet20_augment_result/statistics_plots/bs_128_bs_ori_128_sgd_lrs_[0.05]_pal_mus_[0.1]_ri_100_c_apal_0.9_ori_sgd_lr_0.1/fb_min_step_ratio.pgf"}}
	\end{tabular}
		\vspace{-0.4cm}
	\caption{\textbf{SGD training process with momentum 0.9}. See Figure \ref{la_fig_strategy_metrics} for explanations. The core differences are, that for the proportionality, the noise is higher than in the SGD case. In addition, SGD with momentum  overshoots the locally optimal step size less and does not perform an as exact line search.}
	\label{la_fig_strategy_metrics_momentum}
\end{figure}

%
	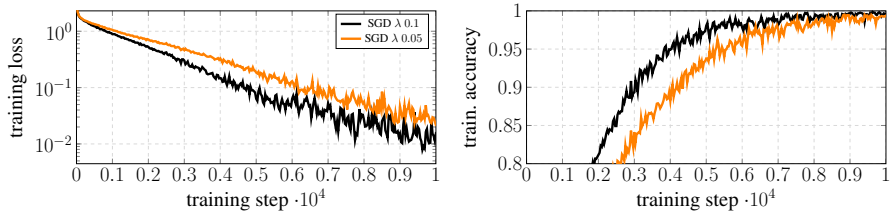
\begin{figure}[h!]
	\newcommand\picscale{0.45}
	\newcommand\lossmin{0}
	\newcommand\lossmax{2.3}
	\newcommand\evalmin{0.5}
	\newcommand\evalmax{1}
	\newcommand\epochs{280}
	\newcommand\picheight{0.5\linewidth}
	\newcommand\legendwidth{9cm}
	\newcommand\legendheight{3cm}
	\centering
	\begin{tabular}{ l r}
		\tikzsetnextfilename{la_best_train_loss}
		\begin{tikzpicture}[scale=\picscale] 
		\begin{axis}[
			width=\linewidth, 
			height=\picheight ,
			grid=major, 
			grid style={dashed,gray!30} , 
			xlabel= training step $\cdot 10^{4}$, 
			xlabel style={font=\LARGE}  ,
			ylabel= training loss,
			ylabel style={font=\LARGE}   ,
			xmin=0,
			xmax=320,
			ymin=\lossmin,ymax=\lossmax,
			x coord trafo/.code={
				\pgflibraryfpuifactive{
					\pgfmathparse{(#1)*(31.25)*(0.0001)}
				}{
					\pgfkeys{/pgf/fpu=true}
					\pgfmathparse{(#1)*(31.25)*(0.0001) }
					\pgfkeys{/pgf/fpu=false}
				}
			},
			ymode=log,
			x tick label style={rotate=0,anchor=near xticklabel,font=\LARGE}   , 
			y tick label style={font=\LARGE},
			p1/.style={draw=black,line width=2pt},
			p2/.style={draw=orange,line width=2pt},
			title style={font=\Huge}   ,
			]
			\addplot [p1] table[x=Step,y=Value,col sep=comma] {line_analysis/figure_data/figure1/train_loss00.csv};
			\addplot [p2] table[x=Step,y=Value,col sep=comma] {line_analysis/figure_data/figure1/train_loss00_0,05.csv};

			\legend{SGD $\lambda$ $0.1$ , SGD  $\lambda$ $0.05$}
		\end{axis}
	\end{tikzpicture}&
	\tikzsetnextfilename{la_best_acc}
	\begin{tikzpicture}[scale=\picscale] 
		\begin{axis}[
			width=\linewidth , 
			height=\picheight,
			grid=major        , 
			grid style={dashed,gray!30}  , 
			xlabel=training step $\cdot 10^{4}$ , 
			xlabel style={font=\LARGE},
			ylabel= train. accuracy,
			ylabel style={font=\LARGE},
			xmin=0,xmax=320,
			ymin=0.8,ymax=1.0,
			scaled x ticks=true,
			x coord trafo/.code={
				\pgflibraryfpuifactive{
					\pgfmathparse{(#1)*(31.25)*(0.0001)}
				}{
					\pgfkeys{/pgf/fpu=true}
					\pgfmathparse{(#1)*(31.25)*(0.0001)}
					\pgfkeys{/pgf/fpu=false}
				}
			},
			x tick label style={rotate=0,anchor=near xticklabel,font=\LARGE}, 
			y tick label style={font=\LARGE},
			p1/.style={draw=black,line width=2pt},
			p2/.style={draw=orange,line width=2pt},
			title style={font=\Huge} ,
			]
			\addplot [p1] table[x=Step,y=Value,col sep=comma] {line_analysis/figure_data/figure1/train_acc00.csv};
			\addplot [p2] table[x=Step,y=Value,col sep=comma] {line_analysis/figure_data/figure1/train_acc00_0,05.csv}; 
		\end{axis}
	\end{tikzpicture}

	\end{tabular}
	\caption{SGD with a locally optimal learning rate of $0.05$ performs worse than SGD with a globally optimal learning rate of $0.01$. Trainings are performed on a ResNet-20 and  8\% of CIFAR-10 with SGD without momentum.\vspace{-1cm}}
	\label{la_fig_sgd_overshoot}
\end{figure}
	
	\vfill
	\clearpage
	\section{Analyses of ResNet-18 and MobileNetV2}
	\label{la_app_resnet18_mobilenet} 
	\subsection{Distance Matrices}

\begin{figure}[h!]
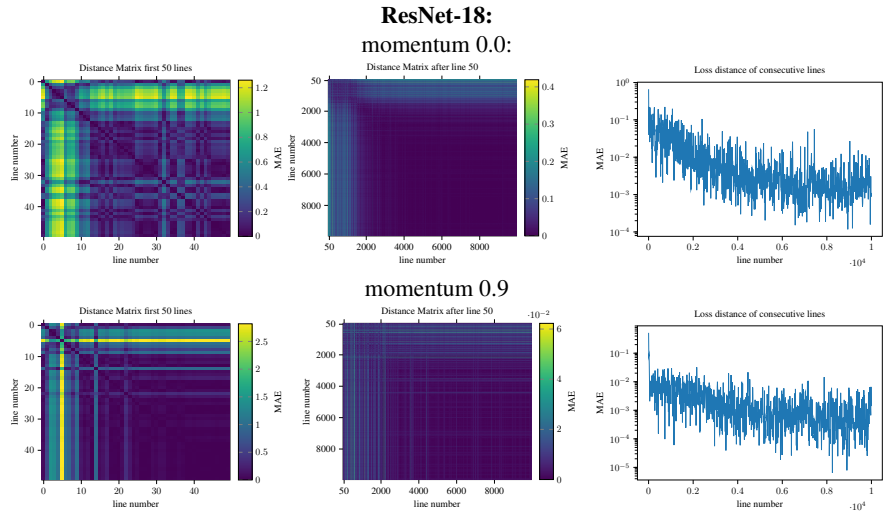

	\centering
	\def\scale{0.365}
	\vspace{-1cm}
	\begin{tabular}{ c c c}
		&\textbf{ResNet-18:}&\\
		&momentum 0.0:&\\
		\tikzsetnextfilename{la_res_net_18_distance_matrix_first_50}
		\scalebox{\scale}{
\begin{tikzpicture}

\begin{axis}[
colorbar,
colorbar style={ylabel={MAE}},
colormap/viridis,
point meta max=1.26129910014931,
point meta min=0,
tick align=outside,
title={Distance Matrix first 50 lines},
x grid style={white!69.0196078431373!black},
xlabel={line number},
xmin=-0.5, xmax=49.5,
xtick pos=both,
xtick style={color=black},
y dir=reverse,
y grid style={white!69.0196078431373!black},
ylabel={line number},
ymin=-0.5, ymax=49.5,
ytick pos=left,
ytick style={color=black}
]
\addplot graphics [includegraphics cmd=\pgfimage,xmin=-0.5, xmax=49.5, ymin=49.5, ymax=-0.5] {\figureTwoDataPath CIFAR10_mom0_resnet18_augment_result/statistics_plots/distance_matrix_first_50-000.png};
\end{axis}

\end{tikzpicture}}&
		\tikzsetnextfilename{la_res_net_18_distance_matrix_after_line50}
		\scalebox{\scale}{\hspace{-0.8cm}
\begin{tikzpicture}

\begin{axis}[
colorbar,
colorbar style={ylabel={MAE}},
colormap/viridis,
point meta max=0.419172447206333,
point meta min=0,
tick align=outside,
title={Distance Matrix after line 50 },
x grid style={white!69.0196078431373!black},
xlabel={line number},
xmin=-0.5, xmax=994.5,
xtick pos=both,
xtick style={color=black},
xtick={5,200,400,600,800},
xticklabels={50,2000,4000,6000,8000},
y dir=reverse,
y grid style={white!69.0196078431373!black},
ylabel={line number},
ymin=-0.5, ymax=994.5,
ytick pos=left,
ytick style={color=black},
ytick={5,200,400,600,800},
yticklabels={50,2000,4000,6000,8000}
]
\addplot graphics [includegraphics cmd=\pgfimage,xmin=-0.5, xmax=994.5, ymin=994.5, ymax=-0.5] {\figureTwoDataPath CIFAR10_mom0_resnet18_augment_result/statistics_plots/distance_matrix_after_line50-000.png};
\end{axis}

\end{tikzpicture}}&
		\tikzsetnextfilename{la_res_net_18_consecutive_line_distances}
		\scalebox{\scale}{\input{"line_analysis/figure_data/line_plots/CIFAR10_mom0_resnet18_augment_result/statistics_plots/consecutive_line_distances.pgf"}}\\
		&momentum 0.9&\\
		\tikzsetnextfilename{la_res_net_18_distance_matrix_first_50_mom}
		\scalebox{\scale}{
\begin{tikzpicture}

\begin{axis}[
colorbar,
colorbar style={ylabel={MAE}},
colormap/viridis,
point meta max=2.81888787694914,
point meta min=0,
tick align=outside,
title={Distance Matrix first 50 lines},
x grid style={white!69.0196078431373!black},
xlabel={line number},
xmin=-0.5, xmax=49.5,
xtick pos=both,
xtick style={color=black},
y dir=reverse,
y grid style={white!69.0196078431373!black},
ylabel={line number},
ymin=-0.5, ymax=49.5,
ytick pos=left,
ytick style={color=black}
]
\addplot graphics [includegraphics cmd=\pgfimage,xmin=-0.5, xmax=49.5, ymin=49.5, ymax=-0.5] {\figureTwoDataPath CIFAR10_mom09_resnet18_augment_result/statistics_plots/distance_matrix_first_50-000.png};
\end{axis}

\end{tikzpicture}}&
		\tikzsetnextfilename{la_res_net_18_distance_matrix_after_line50_mom}
		\scalebox{\scale}{
\begin{tikzpicture}

\begin{axis}[
colorbar,
colorbar style={ylabel={MAE}},
colormap/viridis,
point meta max=0.0622660625968632,
point meta min=0,
tick align=outside,
title={Distance Matrix after line 50 },
x grid style={white!69.0196078431373!black},
xlabel={line number},
xmin=-0.5, xmax=994.5,
xtick pos=both,
xtick style={color=black},
xtick={5,200,400,600,800},
xticklabels={50,2000,4000,6000,8000},
y dir=reverse,
y grid style={white!69.0196078431373!black},
ylabel={line number},
ymin=-0.5, ymax=994.5,
ytick pos=left,
ytick style={color=black},
ytick={5,200,400,600,800},
yticklabels={50,2000,4000,6000,8000}
]
\addplot graphics [includegraphics cmd=\pgfimage,xmin=-0.5, xmax=994.5, ymin=994.5, ymax=-0.5] {\figureTwoDataPath CIFAR10_mom09_resnet18_augment_result/statistics_plots/distance_matrix_after_line50-000.png};
\end{axis}

\end{tikzpicture}}& 
		\tikzsetnextfilename{la_res_net_18_consecutive_line_distances_mom}
		\scalebox{\scale}{\input{"line_analysis/figure_data/line_plots/CIFAR10_mom09_resnet18_augment_result/statistics_plots/consecutive_line_distances.pgf"}}
	\end{tabular}
	\caption{\textbf{ResNet-18:} Distances of the shape of full-batch losses along lines in a window around the current position $s=0$.
		\textbf{Row 1:} SGD without momentum. \textbf{Row 2:} SGD with momentum. Since the offset is not of interest the minimum is shifted to 0 on the y-axis. The distances are rather high for the first 20 lines (left). For the following lines the distances are less than 0.4 MAE (middle) and concentrate around 0.005. The MAEs of the full-batch loss of pairs of consecutive lines are given on the right.}
	\label{la_fig_distance_matrix_resnet18}
\end{figure}
	\begin{figure}[h!]
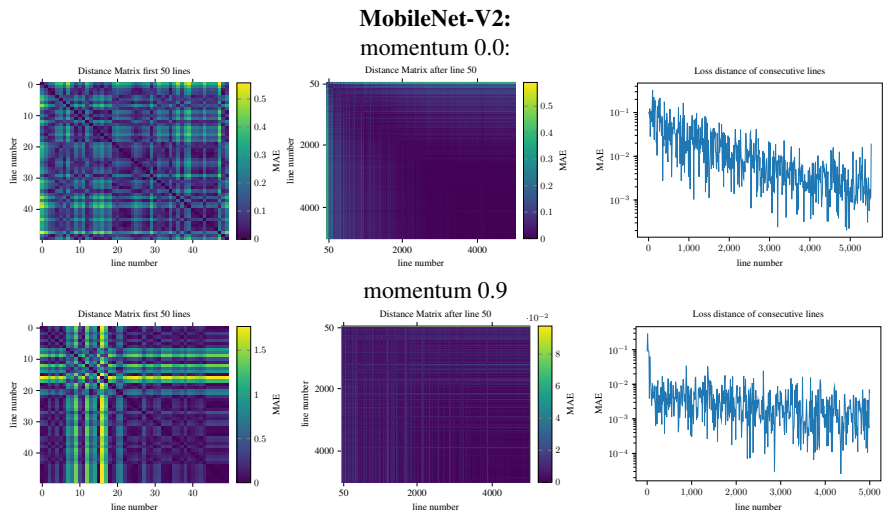

	\centering
	\vspace{-1cm}
	\def\scale{0.365}
	\begin{tabular}{ c c c}
		&\textbf{MobileNet-V2:}&\\
		&momentum 0.0:&\\
		\tikzsetnextfilename{la_mobile_net_distance_matrix_first_50}
		\scalebox{\scale}{
\begin{tikzpicture}

\begin{axis}[
colorbar,
colorbar style={ylabel={MAE}},
colormap/viridis,
point meta max=0.55717556462689,
point meta min=0,
tick align=outside,
title={Distance Matrix first 50 lines},
x grid style={white!69.0196078431373!black},
xlabel={line number},
xmin=-0.5, xmax=49.5,
xtick pos=both,
xtick style={color=black},
y dir=reverse,
y grid style={white!69.0196078431373!black},
ylabel={line number},
ymin=-0.5, ymax=49.5,
ytick pos=left,
ytick style={color=black}
]
\addplot graphics [includegraphics cmd=\pgfimage,xmin=-0.5, xmax=49.5, ymin=49.5, ymax=-0.5] {\figureTwoDataPath CIFAR10_mom0_mobilenet_result/statistics_plots/distance_matrix_first_50-000.png};
\end{axis}

\end{tikzpicture}}&
		\tikzsetnextfilename{la_mobile_net_distance_matrix_after_line50}
		\scalebox{\scale}{\hspace{-0.8cm}
\begin{tikzpicture}

\begin{axis}[
colorbar,
colorbar style={ylabel={MAE}},
colormap/viridis,
point meta max=0.589788390019324,
point meta min=0,
tick align=outside,
title={Distance Matrix after line 50 },
x grid style={white!69.0196078431373!black},
xlabel={line number},
xmin=-0.5, xmax=500,
xtick pos=both,
xtick style={color=black},
xtick={5,200,400},
xticklabels={50,2000,4000},
y dir=reverse,
y grid style={white!69.0196078431373!black},
ylabel={line number},
ymin=-0.5, ymax=500,
ytick pos=left,
ytick style={color=black},
ytick={5,200,400},
yticklabels={50,2000,4000}
]
\addplot graphics [includegraphics cmd=\pgfimage,xmin=-0.5, xmax=500, ymin=500, ymax=-0.5] {\figureTwoDataPath CIFAR10_mom0_mobilenet_result/statistics_plots/distance_matrix_after_line50-000.png};
\end{axis}

\end{tikzpicture}}&
		\tikzsetnextfilename{la_mobile_net_consecutive_line_distances}
		\scalebox{\scale}{\input{"line_analysis/figure_data/line_plots/CIFAR10_mom0_mobilenet_result/statistics_plots/consecutive_line_distances.pgf"}}\\
		&momentum 0.9&\\
		\tikzsetnextfilename{la_mobile_net_distance_matrix_first_50_mom}
		\scalebox{\scale}{
\begin{tikzpicture}

\begin{axis}[
colorbar,
colorbar style={ylabel={MAE}},
colormap/viridis,
point meta max=1.76997652539997,
point meta min=0,
tick align=outside,
title={Distance Matrix first 50 lines},
x grid style={white!69.0196078431373!black},
xlabel={line number},
xmin=-0.5, xmax=49.5,
xtick pos=both,
xtick style={color=black},
y dir=reverse,
y grid style={white!69.0196078431373!black},
ylabel={line number},
ymin=-0.5, ymax=49.5,
ytick pos=left,
ytick style={color=black}
]
\addplot graphics [includegraphics cmd=\pgfimage,xmin=-0.5, xmax=49.5, ymin=49.5, ymax=-0.5] {\figureTwoDataPath CIFAR10_mom09_mobilenet_result/statistics_plots/distance_matrix_first_50-000.png};
\end{axis}

\end{tikzpicture}}&
		\tikzsetnextfilename{la_mobile_net_distance_matrix_after_line50_mom}
		\scalebox{\scale}{
\begin{tikzpicture}

\begin{axis}[
colorbar,
colorbar style={ylabel={MAE}},
colormap/viridis,
point meta max=0.0975745892056773,
point meta min=0,
tick align=outside,
title={Distance Matrix after line 50 },
x grid style={white!69.0196078431373!black},
xlabel={line number},
xmin=-0.5, xmax=500,
xtick pos=both,
xtick style={color=black},
xtick={5,200,400},
xticklabels={50,2000,4000},
y dir=reverse,
y grid style={white!69.0196078431373!black},
ylabel={line number},
ymin=-0.5, ymax=500,
ytick pos=left,
ytick style={color=black},
ytick={5,200,400},
yticklabels={50,2000,4000}
]
\addplot graphics [includegraphics cmd=\pgfimage,xmin=-0.5, xmax=500, ymin=500, ymax=-0.5] {\figureTwoDataPath CIFAR10_mom09_mobilenet_result/statistics_plots/distance_matrix_after_line50-000.png};
\end{axis}

\end{tikzpicture}}& 
		\tikzsetnextfilename{la_mobile_net_consecutive_line_distances_mom}
		\scalebox{\scale}{\input{"line_analysis/figure_data/line_plots/CIFAR10_mom09_mobilenet_result/statistics_plots/consecutive_line_distances.pgf"}}
	\end{tabular}
	\caption{\textbf{MobileNet-V2:} see Figure \ref{la_fig_distance_matrix_resnet18} for explanations. The distances are rather high for the first 25 lines (left). For the following lines the distances are less then 0.6 MAE (middle) and concentrate around 0.01.\vspace{-1.6cm}}
	\label{la_fig_distance_matrix_resnet18_2}
\end{figure}
	\FloatBarrier
	\subsection{Parabolic approximation}
	
\begin{figure}[h!]
	\centering
	\vspace{-0.5cm}
	\def\scale{0.37}
	\begin{tabular}{ c c c}
		&\textbf{ResNet-18:}&\\
		&momentum 0.0:&\\
		\tikzsetnextfilename{la_res_net_18_mae_of_polynomial_fit_of_degree_1}
		\scalebox{\scale}{\input{"line_analysis/figure_data/line_plots/CIFAR10_mom0_resnet18_augment_result/statistics_plots/mae_of_polynomial_fit_of_degree_1.pgf"}}&
		\tikzsetnextfilename{la_res_net_18_mae_of_polynomial_fit_of_degree_2}
		\scalebox{\scale}{\input{"line_analysis/figure_data/line_plots/CIFAR10_mom0_resnet18_augment_result/statistics_plots/mae_of_polynomial_fit_of_degree_2.pgf"}}&
		\tikzsetnextfilename{la_res_net_18_coefficients_of_polynomial_fit_of_degree_2}
		\scalebox{\scale}{\input{"line_analysis/figure_data/line_plots/CIFAR10_mom0_resnet18_augment_result/statistics_plots/coefficients_of_polynomial_fit_of_degree_2.pgf"}}\\
		&momentum 0.9:&\\
		\tikzsetnextfilename{la_res_net_18_mae_of_polynomial_fit_of_degree_1_mom}
		\scalebox{\scale}{\input{"line_analysis/figure_data/line_plots/CIFAR10_mom09_resnet18_augment_result/statistics_plots/mae_of_polynomial_fit_of_degree_1.pgf"}}&
		\tikzsetnextfilename{la_res_net_18_mae_of_polynomial_fit_of_degree_2_mom}
		\scalebox{\scale}{\input{"line_analysis/figure_data/line_plots/CIFAR10_mom09_resnet18_augment_result/statistics_plots/mae_of_polynomial_fit_of_degree_2.pgf"}}&
		\tikzsetnextfilename{la_res_net_18_coefficients_of_polynomial_fit_of_degree_2_mom}
		\scalebox{\scale}{\input{"line_analysis/figure_data/line_plots/CIFAR10_mom09_resnet18_augment_result/statistics_plots/coefficients_of_polynomial_fit_of_degree_2.pgf"}}
	\end{tabular}
	\caption{\textbf{ResNet-18: }MAE of polynomial approximations of the full-batch loss of degree 1 and 2. \textbf{Row 1:} SGD without momentum. \textbf{Row 2:} SGD with momentum. Full-batch losses along lines can be well fitted by polynomials of degree 2. The slope of the approximation stays roughly constant whereas the curvature decreases.}
	\label{la_fig_resnet_18_polynomial_approximations}
\end{figure}

\begin{figure}[h!]
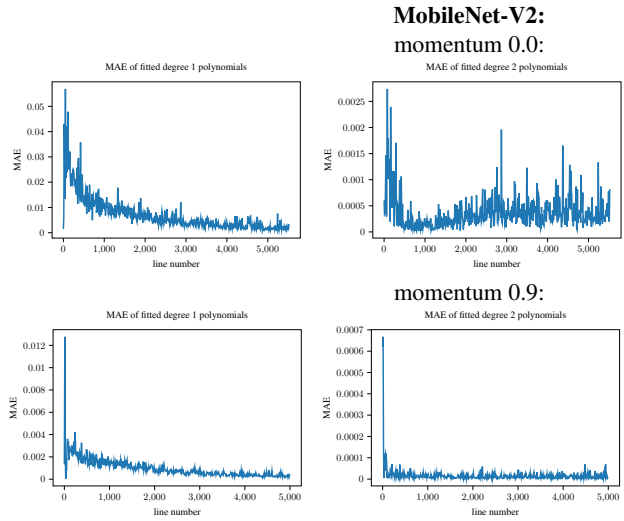

	\centering
	\def\scale{0.37}
	\begin{tabular}{ c c c}
		&\textbf{MobileNet-V2:}&\\
		&momentum 0.0:&\\
		\tikzsetnextfilename{la_mobile_net_mae_of_polynomial_fit_of_degree_1}
		\scalebox{\scale}{\input{"line_analysis/figure_data/line_plots/CIFAR10_mom0_mobilenet_result/statistics_plots/mae_of_polynomial_fit_of_degree_1.pgf"}}&
		\tikzsetnextfilename{la_mobile_net_mae_of_polynomial_fit_of_degree_2}
		\scalebox{\scale}{\input{"line_analysis/figure_data/line_plots/CIFAR10_mom0_mobilenet_result/statistics_plots/mae_of_polynomial_fit_of_degree_2.pgf"}}\\
		&momentum 0.9:&\\
		\tikzsetnextfilename{la_mobile_net_mae_of_polynomial_fit_of_degree_1_mom}
		\scalebox{\scale}{\input{"line_analysis/figure_data/line_plots/CIFAR10_mom09_mobilenet_result/statistics_plots/mae_of_polynomial_fit_of_degree_1.pgf"}}&
		\tikzsetnextfilename{la_mobile_net_mae_of_polynomial_fit_of_degree_2_mom}
		\scalebox{\scale}{\input{"line_analysis/figure_data/line_plots/CIFAR10_mom09_mobilenet_result/statistics_plots/mae_of_polynomial_fit_of_degree_2.pgf"}}
	\end{tabular}
	\caption{\textbf{MobileNet-V2:} for explanations and interpretations see Figure \ref{la_fig_resnet_18_polynomial_approximations} \vspace{-1cm}}
	\label{la_fig_mobilenet_polynomial_approximations}
\end{figure}

	\FloatBarrier
	\subsection{Optimization strategy metrics}
	\begin{figure}[h!]
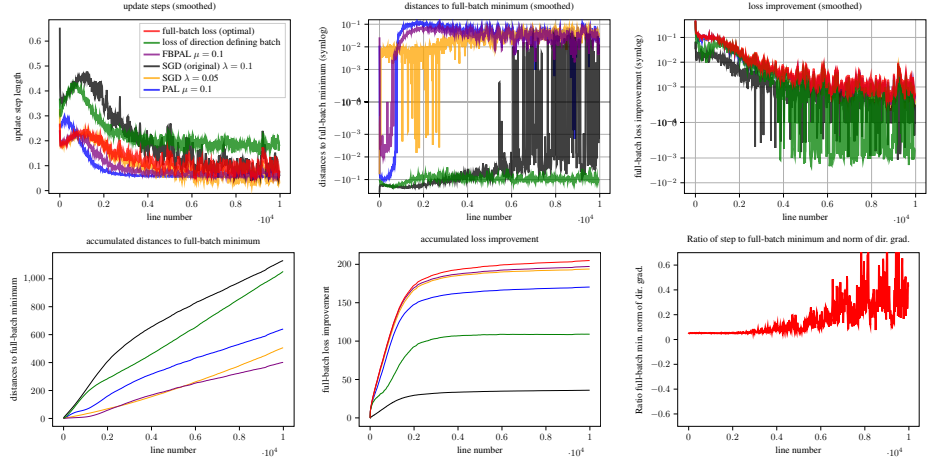

	\centering
	\def\scale{0.36}
	\def\basis{10e-6}
	\vspace{-1cm}
	\begin{tabular}{ c c c }
		\tikzsetnextfilename{la_res_net_18_strategy_update_steps}
		\scalebox{\scale}{\input{"line_analysis/figure_data/line_plots/CIFAR10_mom0_resnet18_augment_result/statistics_plots/bs_128_bs_ori_128_sgd_lrs_[0.05]_pal_mus_[0.1]_ri_100_c_apal_0.9_ori_sgd_lr_0.1/step_step.pgf"}}&
		\pgfplotsset
		{
			y coord trafo/.code={\pgfmathparseFPU{symlog(#1,\basis)}},
			y coord inv trafo/.code={\pgfmathparseFPU{symexp(#1,\basis)}},
		}
		\tikzsetnextfilename{la_res_net_18_strategy_metrics_distances}
		\scalebox{\scale}{\input{"line_analysis/figure_data/line_plots//CIFAR10_mom0_resnet18_augment_result/statistics_plots/bs_128_bs_ori_128_sgd_lrs_[0.05]_pal_mus_[0.1]_ri_100_c_apal_0.9_ori_sgd_lr_0.1/distances_step.pgf"}}&
		\pgfplotsset
		{
			y coord trafo/.code={\pgfmathparseFPU{symlog(#1,\basis)}},
			y coord inv trafo/.code={\pgfmathparseFPU{symexp(#1,\basis)}},
		}
		\tikzsetnextfilename{la_res_net_18_strategy_metrics_improvements}
		\scalebox{\scale}{\input{"line_analysis/figure_data/line_plots//CIFAR10_mom0_resnet18_augment_result/statistics_plots/bs_128_bs_ori_128_sgd_lrs_[0.05]_pal_mus_[0.1]_ri_100_c_apal_0.9_ori_sgd_lr_0.1/improvements_step.pgf"}}\\
		\tikzsetnextfilename{la_res_net_18_strategy_metrics_distances_accumulated}
		\scalebox{\scale}{\input{"line_analysis/figure_data/line_plots//CIFAR10_mom0_resnet18_augment_result/statistics_plots/bs_128_bs_ori_128_sgd_lrs_[0.05]_pal_mus_[0.1]_ri_100_c_apal_0.9_ori_sgd_lr_0.1/distances_step_accumulated.pgf"}}&
		\tikzsetnextfilename{la_res_net_18_strategy_metrics_improvements_accumulated}
		\scalebox{\scale}{\input{"line_analysis/figure_data/line_plots/CIFAR10_mom0_resnet18_augment_result/statistics_plots/bs_128_bs_ori_128_sgd_lrs_[0.05]_pal_mus_[0.1]_ri_100_c_apal_0.9_ori_sgd_lr_0.1/improvements_step_accumulated.pgf"}}&
		\tikzsetnextfilename{la_res_net_18_strategy_metrics_fb_min_mb_grad_ratio}
		\scalebox{\scale}{\input{"line_analysis/figure_data/line_plots/CIFAR10_mom0_resnet18_augment_result/statistics_plots/bs_128_bs_ori_128_sgd_lrs_[0.05]_pal_mus_[0.1]_ri_100_c_apal_0.9_ori_sgd_lr_0.1/fb_min_step_ratio.pgf"}}
	\end{tabular}
	\caption{\textbf{SGD training process without momentum on ResNet18}. Several metrics to compare update step strategies: 1. the performed update steps, 2. the distance to the minimum of the full batch loss ($s_{opt}-s_{upd}$), which is the optimal update step from a local perspective. 3. the loss improvement per step given  as:  $l(0)-l(s_{upd}))$ where $s_{upd}$ is the update step of a strategy. Average smoothing with a kernel size of 25 is applied. In this case the ratio of the full batch minimum location with the norm of the norm of the direction defining gradient increases during the end of the training. The proportionality is only given in the beginning of the training.  }
	\label{la_fig_strategy_metrics_res_net18}
\end{figure}

%
	\begin{figure}[h!]
	\centering
	\def\scale{0.362}
	\def\basis{10e-5}
	\vspace{-1cm}

	\begin{tabular}{ c c c }
		&\textbf{MobileNet-V2 momentum 0}&\\
		\tikzsetnextfilename{la_mobile_net_strategy_update_steps}
		\scalebox{\scale}{\input{"line_analysis/figure_data/line_plots/CIFAR10_mom0_mobilenet_result/statistics_plots/bs_128_bs_ori_128_sgd_lrs_[0.05]_pal_mus_[0.1]_ri_100_c_apal_0.9_ori_sgd_lr_0.1/step_step.pgf"}}&\hspace{-0.25cm}
		\pgfplotsset
		{
			y coord trafo/.code={\pgfmathparseFPU{symlog(#1,\basis)}},
			y coord inv trafo/.code={\pgfmathparseFPU{symexp(#1,\basis)}},
		}
		\tikzsetnextfilename{la_mobile_net_strategy_metrics_distances}
		\scalebox{\scale}{\input{"line_analysis/figure_data/line_plots/CIFAR10_mom0_mobilenet_result/statistics_plots/bs_128_bs_ori_128_sgd_lrs_[0.05]_pal_mus_[0.1]_ri_100_c_apal_0.9_ori_sgd_lr_0.1/distances_step.pgf"}}&\hspace{-0.25cm}
		\pgfplotsset
		{
			y coord trafo/.code={\pgfmathparseFPU{symlog(#1,\basis)}},
			y coord inv trafo/.code={\pgfmathparseFPU{symexp(#1,\basis)}},
		}
		\tikzsetnextfilename{la_mobile_net_strategy_metrics_improvements}
		\scalebox{\scale}{\input{"line_analysis/figure_data/line_plots/CIFAR10_mom0_mobilenet_result/statistics_plots/bs_128_bs_ori_128_sgd_lrs_[0.05]_pal_mus_[0.1]_ri_100_c_apal_0.9_ori_sgd_lr_0.1/improvements_step.pgf"}}\\
		\tikzsetnextfilename{la_mobile_net_strategy_metrics_distances_accumulated}
		\scalebox{\scale}{\input{"line_analysis/figure_data/line_plots/CIFAR10_mom0_mobilenet_result/statistics_plots/bs_128_bs_ori_128_sgd_lrs_[0.05]_pal_mus_[0.1]_ri_100_c_apal_0.9_ori_sgd_lr_0.1/distances_step_accumulated.pgf"}}&\hspace{-0.25cm}
		\tikzsetnextfilename{la_mobile_net_strategy_metrics_improvements_accumulated}
		\scalebox{\scale}{\input{"line_analysis/figure_data/line_plots/CIFAR10_mom0_mobilenet_result/statistics_plots/bs_128_bs_ori_128_sgd_lrs_[0.05]_pal_mus_[0.1]_ri_100_c_apal_0.9_ori_sgd_lr_0.1/improvements_step_accumulated.pgf"}}&\hspace{-0.25cm}
		\tikzsetnextfilename{la_mobile_net_strategy_metrics_fb_min_mb_grad_ratio}
		\scalebox{\scale}{\input{"line_analysis/figure_data/line_plots/CIFAR10_mom0_mobilenet_result/statistics_plots/bs_128_bs_ori_128_sgd_lrs_[0.05]_pal_mus_[0.1]_ri_100_c_apal_0.9_ori_sgd_lr_0.1/fb_min_step_ratio.pgf"}}
	\end{tabular}
	\caption{\textbf{MobileNet-V2 momentum 0}. See Figure \ref{la_fig_strategy_metrics_res_net18} for explanations. \vspace{-1cm}}
		\label{la_fig_strategy_metrics_res_net18_2}
\end{figure}

%
	\begin{figure}[h!]
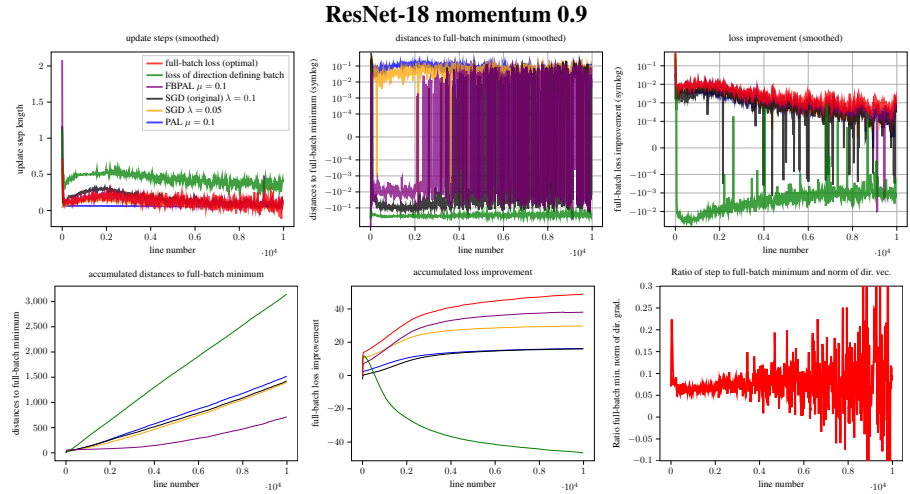

	\centering
	\def\scale{0.362}
	\def\basis{10e-6}
	\begin{tabular}{ c c c }
		&\textbf{ResNet-18 momentum 0.9}&\\
		\tikzsetnextfilename{la_res_net_18_strategy_update_steps_mom}
		\scalebox{\scale}{\input{"line_analysis/figure_data/line_plots/CIFAR10_mom09_resnet18_augment_result/statistics_plots/bs_128_bs_ori_128_sgd_lrs_[0.05]_pal_mus_[0.1]_ri_100_c_apal_0.9_ori_sgd_lr_0.1/step_step.pgf"}}&\hspace{-0.25cm}
		\pgfplotsset
		{
			y coord trafo/.code={\pgfmathparseFPU{symlog(#1,\basis)}},
			y coord inv trafo/.code={\pgfmathparseFPU{symexp(#1,\basis)}},
		}
		\tikzsetnextfilename{la_res_net_18_strategy_metrics_distances_mom}
		\scalebox{\scale}{\input{"line_analysis/figure_data/line_plots/CIFAR10_mom09_resnet18_augment_result/statistics_plots/bs_128_bs_ori_128_sgd_lrs_[0.05]_pal_mus_[0.1]_ri_100_c_apal_0.9_ori_sgd_lr_0.1/distances_step.pgf"}}&\hspace{-0.25cm}
		\pgfplotsset
		{
			y coord trafo/.code={\pgfmathparseFPU{symlog(#1,\basis)}},
			y coord inv trafo/.code={\pgfmathparseFPU{symexp(#1,\basis)}},
		}
		\tikzsetnextfilename{la_res_net_18_strategy_metrics_improvements_mom}
		\scalebox{\scale}{\input{"line_analysis/figure_data/line_plots/CIFAR10_mom09_resnet18_augment_result/statistics_plots/bs_128_bs_ori_128_sgd_lrs_[0.05]_pal_mus_[0.1]_ri_100_c_apal_0.9_ori_sgd_lr_0.1/improvements_step.pgf"}}\\
		\tikzsetnextfilename{la_res_net_18_strategy_metrics_distances_accumulated_mom}
		\scalebox{\scale}{\input{"line_analysis/figure_data/line_plots/CIFAR10_mom09_resnet18_augment_result/statistics_plots/bs_128_bs_ori_128_sgd_lrs_[0.05]_pal_mus_[0.1]_ri_100_c_apal_0.9_ori_sgd_lr_0.1/distances_step_accumulated.pgf"}}&\hspace{-0.25cm}
		\tikzsetnextfilename{la_res_net_18_strategy_metrics_improvements_accumulated_mom}
		\scalebox{\scale}{\input{"line_analysis/figure_data/line_plots/CIFAR10_mom09_resnet18_augment_result/statistics_plots/bs_128_bs_ori_128_sgd_lrs_[0.05]_pal_mus_[0.1]_ri_100_c_apal_0.9_ori_sgd_lr_0.1/improvements_step_accumulated.pgf"}}&\hspace{-0.25cm}
		\tikzsetnextfilename{la_res_net_18_strategy_metrics_fb_min_mb_grad_ratio_mom}
		\scalebox{\scale}{\input{"line_analysis/figure_data/line_plots/CIFAR10_mom09_resnet18_augment_result/statistics_plots/bs_128_bs_ori_128_sgd_lrs_[0.05]_pal_mus_[0.1]_ri_100_c_apal_0.9_ori_sgd_lr_0.1/fb_min_step_ratio.pgf"}}
	\end{tabular}
	
	\caption{\textbf{ResNet-18 momentum 0.9}. See Figure \ref{la_fig_strategy_metrics_res_net18} for explanations. In the case of momentum SGD is not able to perform such an exact line search as in the case without monemtum since the norm of the momentum vector is not directly related to the loss of the current line considered.}
	\label{la_fig_strategy_metrics_res_net18_3}
\end{figure}

%
	\begin{figure}[h!]
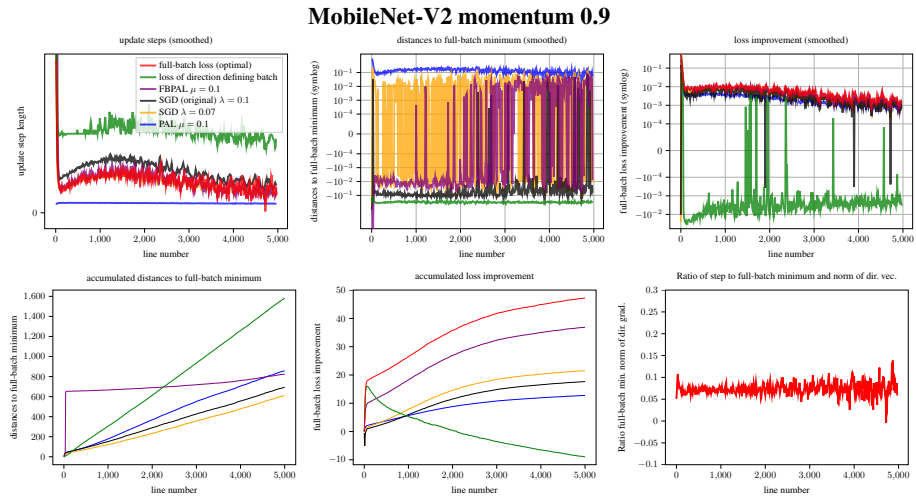

	\centering
	\def\scale{0.362}
	\def\basis{10e-6}
	
	\begin{tabular}{ c c c }
		&\textbf{MobileNet-V2 momentum 0.9}&\\
		\tikzsetnextfilename{la_mobile_net_strategy_update_steps_mom}
		\scalebox{\scale}{\input{"line_analysis/figure_data/line_plots/CIFAR10_mom09_mobilenet_result/statistics_plots/bs_128_bs_ori_128_sgd_lrs_[0.07]_pal_mus_[0.1]_ri_100_c_apal_0.9_ori_sgd_lr_0.1/step_step.pgf"}}&\hspace{-0.25cm}
		\pgfplotsset
		{
			y coord trafo/.code={\pgfmathparseFPU{symlog(#1,\basis)}},
			y coord inv trafo/.code={\pgfmathparseFPU{symexp(#1,\basis)}},
		}
		\tikzsetnextfilename{la_mobile_net_strategy_metrics_distances_mom}
		\scalebox{\scale}{\input{"line_analysis/figure_data/line_plots/CIFAR10_mom09_mobilenet_result/statistics_plots/bs_128_bs_ori_128_sgd_lrs_[0.07]_pal_mus_[0.1]_ri_100_c_apal_0.9_ori_sgd_lr_0.1/distances_step.pgf"}}&\hspace{-0.25cm}
		\pgfplotsset
		{
			y coord trafo/.code={\pgfmathparseFPU{symlog(#1,\basis)}},
			y coord inv trafo/.code={\pgfmathparseFPU{symexp(#1,\basis)}},
		}
		\tikzsetnextfilename{la_mobile_net_strategy_metrics_improvements_mom}
		\scalebox{\scale}{\input{"line_analysis/figure_data/line_plots/CIFAR10_mom09_mobilenet_result/statistics_plots/bs_128_bs_ori_128_sgd_lrs_[0.07]_pal_mus_[0.1]_ri_100_c_apal_0.9_ori_sgd_lr_0.1/improvements_step.pgf"}}\\
		\tikzsetnextfilename{la_mobile_net_strategy_metrics_distances_accumulated_mom}
		\scalebox{\scale}{\input{"line_analysis/figure_data/line_plots/CIFAR10_mom09_mobilenet_result/statistics_plots/bs_128_bs_ori_128_sgd_lrs_[0.07]_pal_mus_[0.1]_ri_100_c_apal_0.9_ori_sgd_lr_0.1/distances_step_accumulated.pgf"}}&\hspace{-0.25cm}
		\tikzsetnextfilename{la_mobile_net_strategy_metrics_improvements_accumulated_mom}
		\scalebox{\scale}{\input{"line_analysis/figure_data/line_plots/CIFAR10_mom09_mobilenet_result/statistics_plots/bs_128_bs_ori_128_sgd_lrs_[0.07]_pal_mus_[0.1]_ri_100_c_apal_0.9_ori_sgd_lr_0.1/improvements_step_accumulated.pgf"}}&\hspace{-0.25cm}
		\tikzsetnextfilename{la_mobile_net_strategy_metrics_fb_min_mb_grad_ratio_mom}
		\scalebox{\scale}{\input{"line_analysis/figure_data/line_plots/CIFAR10_mom09_mobilenet_result/statistics_plots/bs_128_bs_ori_128_sgd_lrs_[0.07]_pal_mus_[0.1]_ri_100_c_apal_0.9_ori_sgd_lr_0.1/fb_min_step_ratio.pgf"}}
	\end{tabular}
	
	\caption{\textbf{MobileNet-V2 momentum 0.9}. See Figure \ref{la_fig_strategy_metrics_res_net18} and 7 for explanations and interpretations.}
	\label{la_fig_strategy_metrics_res_net18_4}
\end{figure}

%
	\FloatBarrier
	\vfill
	\clearpage
	\subsection{Batch size comparison}
	\begin{figure}[h!]
	\centering
	\def\scale{0.36}
	\vspace{-1.0cm}
	\begin{tabular}{ c c}
		\textbf{ResNet-18 momentum 0}&\\
		\tikzsetnextfilename{la_res_net_18_batch_size_sgd_improvements_step_accumulated}
		\scalebox{\scale}{\input{"line_analysis/figure_data/line_plots//CIFAR10_mom0_resnet18_augment_result/statistics_plots/batch_size_comparison/SGDoriginallambda=0,1_improvements_step_accumulated.pgf"}}&\hspace{0.1cm}
		\tikzsetnextfilename{la_res_net_18_batch_size_pal_improvements_step_accumulated}	
		\scalebox{\scale}{\input{"line_analysis/figure_data/line_plots/CIFAR10_mom0_resnet18_augment_result/statistics_plots/batch_size_comparison/PALmu=0,1_improvements_step_accumulated.pgf"}}\\
		\tikzsetnextfilename{la_res_net_18_batch_size_directional_derivative}
		\scalebox{\scale}{\input{"line_analysis/figure_data/line_plots/CIFAR10_mom0_resnet18_augment_result/statistics_plots/batch_size_comparison/directional_derivatives.pgf"}}&
		\tikzsetnextfilename{la_res_net_18_batch_size_directional_derivatives_ratio}
		\scalebox{\scale}{\input{"line_analysis/figure_data/line_plots/CIFAR10_mom0_resnet18_augment_result/statistics_plots/batch_size_comparison/directional_derivatives_ratio.pgf"}}
	\end{tabular}\\
	\caption{\textbf{Row 1} Comparing the influence of the batch size on the loss improvement. Left: SGD with the original learning rate of 0.1. Right: parabolic approximation line search (PAL).
		\textbf{Row 2: } Analysis of the relation of the batch size to the absolute directional derivative (=gradient norm) which shows in detail that increasing the batch size has a similar effect as decreasing the learning rate by the same factor.  \textbf{e.r.}  stands for expected ratio.}.
	\label{la_fig_distances_batchwise_app}
\end{figure}

	\begin{figure}[b!]
	\centering
	\vspace{-1.7cm}
	\def\scale{0.36}
	\begin{tabular}{ c c}
		\textbf{MobileNet-V2 momentum 0}&\\
		\tikzsetnextfilename{la_mobile_net_batch_size_sgd_improvements_step_accumulated}
		\scalebox{\scale}{\input{"line_analysis/figure_data/line_plots//CIFAR10_mom0_mobilenet_result/statistics_plots/batch_size_comparison/SGDoriginallambda=0,1_improvements_step_accumulated.pgf"}}&\hspace{0.1cm}
		\tikzsetnextfilename{la_mobile_net_batch_size_pal_improvements_step_accumulated}	
		\scalebox{\scale}{\input{"line_analysis/figure_data/line_plots/CIFAR10_mom0_mobilenet_result/statistics_plots/batch_size_comparison/PALmu=0,1_improvements_step_accumulated.pgf"}}\\
		\tikzsetnextfilename{la_mobile_net_batch_size_directional_derivative}
		\scalebox{\scale}{\input{"line_analysis/figure_data/line_plots/CIFAR10_mom0_mobilenet_result/statistics_plots/batch_size_comparison/directional_derivatives.pgf"}}&
		\tikzsetnextfilename{la_mobile_net_batch_size_directional_derivatives_ratio}
		\scalebox{\scale}{\input{"line_analysis/figure_data/line_plots/CIFAR10_mom0_mobilenet_result/statistics_plots/batch_size_comparison/directional_derivatives_ratio.pgf"}}
	\end{tabular}\\
	\caption{\textbf{MobileNet-V2 momentum 0:} See Figure \ref{la_fig_distances_batchwise_app} for explanations. }.
	\label{la_fig_distances_batchwise_app2}
\end{figure}

%
%
%

%
%
%
%

%
\end{document}